\documentclass{article}

\usepackage[a4paper,
left=3cm,
right=3cm,
top=3cm,
bottom=3cm]{geometry}

\usepackage[utf8]{inputenc}
\usepackage[T1]{fontenc}
\usepackage{hyperref}
\usepackage{url}
\usepackage{booktabs}
\usepackage{amsfonts}
\usepackage{nicefrac}
\usepackage{xcolor}
\usepackage{graphicx} 
\usepackage{svg}
\usepackage{amsmath}
\usepackage{subcaption}
\usepackage{float}

\begin{document}
\title{DSS-GAN: Directional State Space GAN with Mamba backbone for Class-Conditional Image Synthesis}
\author{
Aleksander Ogonowski$^{1,2}$\thanks{Corresponding author: \texttt{aleksander.ogonowski.dokt@pw.edu.pl}, \texttt{aleksander.ogonowski@ncbj.gov.pl}},
Konrad Klimaszewski$^{2}$,
Przemysław Rokita$^{1}$ \\
[0.5em]
$^{1}$Warsaw University of Technology \\
$^{2}$National Centre for Nuclear Research
}

\maketitle

\begin{abstract}
We present DSS-GAN, the first generative adversarial network to employ Mamba as a hierarchical
generator backbone for noise-to-image synthesis.
The central contribution is Directional Latent Routing (DLR), a novel
conditioning mechanism that decomposes the latent vector into
direction-specific subvectors, each jointly projected with a class embedding
to produce a feature-wise affine modulation of the corresponding Mamba scan.
Unlike conventional class conditioning that injects a global signal,
DLR couples class identity and latent structure along distinct spatial axes
of the feature map, applied consistently across all generative scales.
DSS-GAN achieves improved FID, KID, and precision-recall scores compared
to StyleGAN2-ADA~\cite{karras2020stylegan2} across multiple tested datasets.
Analysis of the latent space reveals that directional subvectors exhibit
measurable specialization: perturbations along individual components
produce structured, direction-correlated changes in the synthesized image.
\end{abstract}

\section{Introduction}

Generative image modeling has undergone a fundamental shift in recent years.
Diffusion models~\cite{ho2020ddpm, rombach2022ldm} and autoregressive
approaches~\cite{vanoord2017vqvae, esser2021vqgan} now dominate the field,
delivering impressive sample quality at the cost of iterative inference
passes~\cite{song2021score, dhariwal2021diffusion} and latent spaces that
are notoriously difficult to interpret and control~\cite{kwon2022diffusion_latent}.
GANs, however, remain the natural choice wherever single-pass inference
speed matters~\cite{sauer2022styleganxl, kang2023gigagan} and where
direct, deterministic control over the generated output is
required~\cite{goodfellow2014gan, karras2019stylegan, karras2020stylegan2,
brock2019biggan}~- in particular for applications demanding precise
class- and structure-conditioned synthesis~\cite{mirza2014cgan,
odena2017acgan, miyato2018projection}.

One of the central challenges in conditional GANs is how to efficiently
inject class and structure information across spatial scales while
preserving long-range
dependencies~\cite{brock2019biggan, karras2020stylegan2, bourou2024survey}.
Convolutional generators capture local features well, but their receptive
field is inherently bounded~\cite{lecun1998cnn}.
ViT-based generators~\cite{esser2021vqgan, kang2023gigagan} recover global
context at the price of $\mathcal{O}(N^2)$ attention complexity~--- at
$256{\times}256$ resolution this amounts to $65\,536$ tokens and
prohibitive memory requirements.
State space models (SSMs)~\cite{gu2022s4, gu2023mamba}, and Mamba in
particular, offer linear-complexity sequence modeling while retaining
long-range dependencies~--- initially demonstrated in discriminative
tasks such as classification and
segmentation~\cite{liu2024vmamba, zhu2024vim, nguyen2024mamband}.
Their application to generative modeling has since expanded, yet remains
confined to image-to-image translation and
restoration~\cite{atli2024i2imamba, jin2025hmsmambagan, wu2025mambaragan,
hong2026stymam}, where Mamba acts as an encoder or bottleneck module
processing existing image features, and to diffusion
frameworks~\cite{teng2024dim, hu2024zigma, phung2024dimsum}, where it
replaces the denoising backbone but inherits the iterative inference
pipeline.
Their potential as a GAN generator backbone for class-conditioned
noise-to-image synthesis remains unexplored.
We propose \textbf{DSS-GAN} (\emph{Directional State Space GAN})~--- the
first GAN to employ Mamba as a generator backbone for
noise-to-image synthesis. Existing Mamba-GAN works are confined to image
translation and restoration~\cite{wu2025mambaragan, jin2025hmsmambagan},
where Mamba acts as an encoder module~--- generation from noise has remained
the exclusive domain of CNN and ViT architectures.

The core contribution is \textbf{DLR} (\emph{Directional Latent Routing}):
the latent vector $\mathbf{z}$ is decomposed into direction-specific
subvectors, each jointly conditioned with a class embedding to modulate
the token sequence of the corresponding Mamba scan. The same routing is
applied consistently across all scales, providing globally coherent
conditioning.

DSS-GAN achieves comparable or superior FID, KID, and precision-recall scores to
StyleGAN2-ADA~\cite{karras2020stylegan2} across all tested datasets
while requiring more than $3\times$ fewer parameters.
Latent space analysis demonstrates that directional subvectors exert
distinct, direction-correlated influence on class-specific spatial features
of the synthesized image.

The source code is publicly available at \url{https://github.com/dssgan/DSS_GAN}.
\section{Related Works}

\paragraph{Class-conditional image synthesis and latent modulation.}
Early conditional GANs inject class information by concatenating a class
embedding with the noise vector (e.g., cGAN~\cite{mirza2014cgan}) or by
adding an auxiliary classification objective (e.g.,
AC-GAN~\cite{odena2017acgan}).
BigGAN~\cite{brock2019biggan} scales this paradigm to large category sets
using class-conditional batch normalization: a shared class embedding,
concatenated with a chunk of the latent vector, is linearly projected to
per-channel gains and biases $(\gamma, \beta)$ that modulate each
residual block --- a mechanism closely related to Feature-wise Linear
Modulation (FiLM)~\cite{perez2018film}, which formalizes this class of
affine conditioning layers.
Notably, BigGAN also splits $\mathbf{z}$ into per-resolution chunks
(skip-z connections), providing a form of scale-wise latent routing --- but
without any direction-specific decomposition of the conditioning signal.
StyleGAN~\cite{karras2019stylegan} and StyleGAN2~\cite{karras2020stylegan2}
introduce a learned mapping network that transforms the latent vector into
a style code --- a learned intermediate representation that disentangles
the latent space --- which is then used to globally modulate the synthesis 
process at each layer, either via per-channel affine shifts applied to 
feature maps (StyleGAN) or through weight scaling and demodulation (StyleGAN2).
Style mixing allows different codes to be applied at different resolution
stages, providing coarse-to-fine hierarchical control, but within each
layer the conditioning remains spatially uniform --- the same modulation 
is applied to every spatial position of the feature map.
StyleGAN3~\cite{karras2021stylegan3} further reformulates synthesis as
continuous signal processing to eliminate aliasing artifacts, but retains
the same globally-uniform conditioning scheme.
In all these approaches, the latent vector and class signal are either
concatenated or jointly mapped to a single global conditioning vector, with
no mechanism to assign different components of the latent space to distinct
spatial structures of the image.

SPADE~\cite{park2019spade} breaks the spatial uniformity constraint by
conditioning on a dense semantic map to produce spatially-varying affine
shifts, achieving fine-grained local control --- but at the cost of requiring
an explicit pixel-aligned spatial input rather than a compact latent code.
InfoGAN~\cite{chen2016infogan} takes a different approach to latent
structure: it decomposes $\mathbf{z}$ into interpretable components by
maximizing mutual information between latent variables and generated
observations, achieving unsupervised disentanglement through a training
objective rather than by architectural design.
ViTGAN~\cite{lee2021vitgan}, a representative ViT-based generator, adopts
self-modulated layer normalization (SLN), an AdaIN-style mechanism in which
the latent code globally modulates transformer token representations via
learned affine parameters --- without class conditioning or
direction-specific differentiation of the latent signal.
The DLR mechanism in DSS-GAN differs from all of the above: it couples
class identity and latent structure along distinct scan directions of the
feature map by architectural design, without requiring a spatial input map,
a mutual information objective, or quadratic attention.

\paragraph{State space models in vision and generative modeling.}
Mamba~\cite{gu2023mamba} introduces selective state space models with
linear complexity, enabling efficient modeling of long-range dependencies.
Follow-up works adapt Mamba to vision tasks via multi-directional scanning
strategies: VMamba~\cite{liu2024vmamba} and Vim~\cite{zhu2024vim} achieve
strong results in classification and segmentation, while
Mamba-ND~\cite{nguyen2024mamband} extends scanning to arbitrary
dimensionalities. However, all these works employ Mamba exclusively as an
encoder backbone for discriminative tasks.
In the GAN setting, MambaRAGAN~\cite{wu2025mambaragan} and
HMSMamba-GAN~\cite{jin2025hmsmambagan} incorporate Mamba as an encoder
module for image-to-image translation and restoration --- generation from
noise remains unexplored. 

\paragraph{Mamba in diffusion models.}
Recent work has explored Mamba as a denoising backbone in diffusion models.
DiM~\cite{teng2024dim} replaces the U-Net with a multi-directional Mamba
architecture for efficient high-resolution synthesis; ZigMa~\cite{hu2024zigma}
introduces spatially continuous zigzag scanning schemes to improve
position-awareness; DiMSUM~\cite{phung2024dimsum} integrates wavelet-domain
scanning with spatial Mamba via cross-attention fusion.
In all these works, Mamba processes noisy image tokens to predict noise at
each diffusion step --- a discriminative role within an iterative refinement
process. DSS-GAN is the first GAN to use Mamba as a hierarchical generator
backbone for noise-to-image synthesis.

\section{DSS GAN}
\subsection{Architecture Overview}

\begin{figure}[h]
  \centering
  \includegraphics[width=\linewidth]{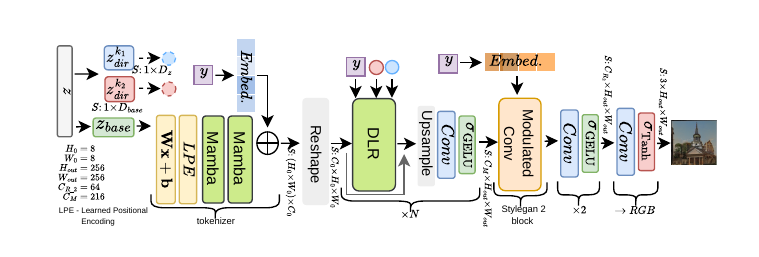}
  \caption{DSS-GAN generator architecture (Only two scan directions shown for
  clarity). $Z_{base}$ is base global latent vector, $z^{k_K}_{dir}$ are directional latent vectors. For $256 \times 256$ resolution $N$ is $5$ and $C_0$ - feature map channels in DLR blocks is $148$. }
  \label{fig:arch}
\end{figure}

DSS-GAN is a class-conditional GAN evaluated at up to $512{\times}512$
resolution. The generator combines a hierarchical Mamba backbone with a
StyleGAN2-inspired convolutional refinement block at the final resolution,
exploiting the complementary strengths of both architectures: Mamba builds
global coherence across scales, while the CNN adds local high-frequency
detail. The overall generator 
architecture is illustrated in Fig.~\ref{fig:arch}.

The discriminator follows the StyleGAN2-ADA
architecture~\cite{karras2020ada}. A ViT-based discriminator was evaluated
early in development but proved destabilizing: jointly optimizing a novel
generator backbone and a non-standard discriminator introduces too many
simultaneous degrees of freedom. We therefore retained the well-validated
StyleGAN2-ADA discriminator to isolate the contribution of the generator.

Both generator and discriminator are trained with the non-saturating GAN loss~\cite{goodfellow2014gan} and R1 gradient penalty~\cite{mescheder2018training}; full training configuration is provided in Appendix~\ref{app:training}.

\paragraph{Latent vector and class conditioning.}
The latent vector $\mathbf{z} \in \mathbb{R}^{D_z}$ is partitioned into two
components: a base latent $\mathbf{z}_\text{base}$, consumed once by the
tokenizer ~\ref{sec:mamba_tokenzer}, and a directional routing vector $\mathbf{z}_\text{dir}$, which is
further split into $K$ equal subvectors - one per scan direction (detailed
in Section~\ref{sec:dlr}). The key property of $\mathbf{z}_\text{dir}$ is
that it is \emph{fixed across all resolution stages}: the same vector is
re-injected independently at every DLR ~\ref{sec:dlr} , so the directional routing
signal remains constant throughout the hierarchy. The class label $y$ is
treated identically 
- re-injected at every DLR stage rather than propagated
through the spatial feature map, which carries only the
progressively refined image representation between blocks.
Table~\ref{tab:arch} summarizes the routing for $512{\times}512$ image generation.

\begin{table}[h]
\centering
\setlength{\tabcolsep}{5pt}
\small
\begin{tabular}{llccc}
\toprule
Block & Resolution &
$\mathbf{z}_\text{base}$ & $\mathbf{z}_\text{dir}$ & $y$ \\
\midrule
Tokenizer  & $8{\times}8$                              & seed & ---    & additive bias \\
DLR        & $8{\times}8 \to 256{\times}256$ (×6)      & ---        & shared, per-direction & shared, per-direction    \\
Conv       & $512{\times}512$                          & ---        & ---    & global                   \\
toRGB      & $512{\times}512$                          & ---        & ---    & ---                      \\
\bottomrule
\end{tabular}
\caption{Generator routing summary for the DSS-GAN blocks: Tokenizer (\ref{sec:mamba_tokenzer}), DLR (\ref{sec:dlr}), Conv and toRGB (\ref{sec:aggregation}). The $\mathbf{z}_\text{base}$ seeds the
tokenizer only; $\mathbf{z}_\text{dir}$ and $y$ are re-injected unchanged at
every DLR stage. }
\label{tab:arch}
\end{table}

\paragraph{Mamba tokenizer.}
\label{sec:mamba_tokenzer}
The $\mathbf{z}_\text{base}$ vector is mapped to an initial token sequence by a learned
linear projection, then processed by two Mamba blocks before being reshaped
into an $8{\times}8$ spatial grid. Each Mamba block projects the input to a
higher-dimensional space, applies the selective SSM with a depthwise
convolutional mixing layer, and projects back to the original dimension, with
a residual connection around the full block. For a detailed description of the Mamba architecture the reader is referred to~\cite{gu2023mamba}.

Crucially, the class label $y$ is injected \emph{after} the Mamba blocks
and \emph{before} the reshape, as a global additive bias
$\mathbf{t} \leftarrow \mathbf{t} + \mathrm{Embed}(y)$ to the token sequence.
This ordering is a deliberate design choice: the SSM establishes token
interactions solely on the basis of $\mathbf{z}_\text{base}$, without the
class signal influencing the recurrent state. The class embedding then shifts
the entire token representation by a class-specific offset. As a result,
intra-class structural diversity is preserved --- samples of the same class
differ in layout and composition, which is determined entirely by
$\mathbf{z}_\text{base}$, and share only a class-level bias in the initial
spatial seed.

\paragraph{DLR blocks and upsampling.}
A sequence of resolution stages processes the feature map
from $8{\times}8$ to $128{\times}128$ (Figure~\ref{fig:arch}). Each stage
applies one or more DLR blocks (Section~\ref{sec:dlr}), each wrapped in a
residual connection. To adapt channel dimensions between stages, a two-step
upsampling is applied: nearest-neighbour interpolation doubles the spatial
resolution, followed by a learned convolutional projection that adjusts the
number of channels. Since the convolution is applied after the interpolation
step, the spatial layout is determined solely by the upsampling and is not
distorted by learned kernel weights. Upsampling via transposed convolutions
is avoided due to known checkerboard artifacts~\cite{odena2016deconv},
following ProGAN~\cite{karras2018progressive}.

\paragraph{Convolutional refinement.}
The final resolution stage uses a StyleGAN2-inspired convolutional block
rather than a Mamba stage. Each additional Mamba stage quadruples the token
count, so replacing the highest-resolution stage with a convolutional block
balances global coherence with local high-frequency refinement that
convolutional layers handle more efficiently.

\begin{table}[h]
\centering
\small
\caption{Ablation: effect of Stylegan2 CNN refinement block boundary resolution on AFHQ (\textit{cat, dog, wild})$128{\times}128$.
All variants share the same training configuration.}
\label{tab:ablation_cnn}
\setlength{\tabcolsep}{5pt}
\begin{tabular}{l cccccc | cccccc}
\toprule
& \multicolumn{6}{c|}{Global} & \multicolumn{6}{c}{Mean Class} \\
\cmidrule(lr){2-7}\cmidrule(lr){8-13}
Stylegan2\\ refinement\\ from & FID $\downarrow$ & KID $\downarrow$ & P $\uparrow$ & R $\uparrow$ & D $\uparrow$ & C $\uparrow$ & FID $\downarrow$ & KID $\downarrow$ & P $\uparrow$ & R $\uparrow$ & D $\uparrow$ & C $\uparrow$ \\
\midrule
$None$
  & 21.75 & 5.26 & .815 & .242 & 1.078 & .577
  & 27.35 & 13.86 & .814 & .188 & 1.031 & .563 \\
$128{\times}128$
  & \textbf{11.66} & \textbf{2.71} & \textbf{.822} & \textbf{.407} & \textbf{1.144} & \textbf{.748}
  & \textbf{12.93} & \textbf{5.67} & \textbf{.826} & \textbf{.394} & \textbf{1.125} & \textbf{.744} \\
$64{\times}64$
  & 18.48 & 8.27 & .574 & .264 & 0.357 & .388
  & 20.75 & 13.45 & .541 & .264 & 0.322 & .364 \\
$32{\times}32$
  & 25.67 & 9.19 & .621 & .038 & 0.394 & .323
  & 29.18 & 16.62 & .596 & .035 & 0.366 & .297 \\
$16{\times}16$
  & 50.30 & 23.61 & .248 & .009 & 0.104 & .098
  & 59.08 & 40.49 & .241 & .005 & 0.100 & .097 \\
\bottomrule
\end{tabular}
\end{table}

Table~\ref{tab:ablation_cnn} reports the effect of progressively replacing
Mamba stages with StyleGAN2-inspired convolutional blocks, evaluated on
AFHQ $128{\times}128$. Removing the convolutional block entirely increases
precision but substantially reduces recall --- the Mamba backbone alone
produces sharp but poorly diverse outputs, collapsing toward
class-conditional means at the finest spatial scales. Conversely, the
convolutional block receives no directional latent input, so extending it
to lower resolutions progressively displaces latent routing, which is
responsible for output diversity --- hence the monotonic recall collapse
as the CNN boundary moves earlier. The convolutional block is most
effective as a purely local refinement stage at the final resolution,
where it adds high-frequency detail without interfering with the latent

\subsection{Directional Latent Routing}
\label{sec:dlr}

Directional Latent Routing (DLR) is the central conditioning mechanism of
DSS-GAN. Rather than injecting a single global class and latent signal into
every layer, DLR decomposes $\mathbf{z}$ into direction-specific subvectors
and couples each with a dedicated class embedding to produce a feature-wise
affine modulation of the corresponding Mamba scan. Intuitively, each scan
direction acts as a distinct perceptual axis, while the direction weighting network plays the role of class-aware attentional gating --- determining how much each axis contributes to the
final representation, conditioned jointly on the class label and the
directional latent.
\begin{figure}[h]
  \centering
  \includegraphics[width=0.8\linewidth]{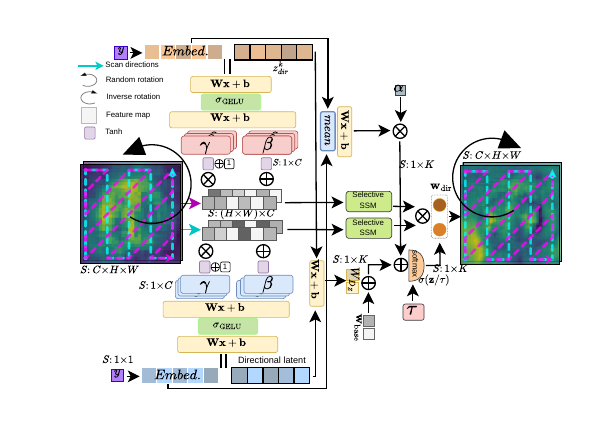}
  \caption{DLR block. For each scan direction $k$, the directional latent
  $\mathbf{z}_\text{dir}^k$ and class embedding $\mathbf{e}_k$ are projected
  to affine parameters $(\boldsymbol{\gamma}_k, \boldsymbol{\beta}_k)$ that
  modulate the token sequence before the Mamba SSM. Both the class and the directional latent determine the contribution of each scan direction to the generated feature map, with directional weights $w^{\text{dir}}_k = f(y, z^{\text{dir}}_k)$ satisfying $\sum_{k=1}^{K} w_{\text{dir}}^k = 1$. A random $180^\circ$ rotation
  is applied before the block and inverted after unscan.}
  \label{fig:dlr}
\end{figure}

\subsubsection{Scan, Unscan, and Rotation}
\label{sec:scan}

\paragraph{Scan directions.} Mamba operates on one-dimensional sequences. To apply it to a
two-dimensional feature map $\mathbf{h} \in \mathbb{R}^{C \times H \times W}$, the spatial grid must first be serialized into a token sequence ---
a \emph{scan} operation --- and the output must be deserialized back to
the spatial grid after the SSM --- an \emph{unscan} operation. The choice
of serialization order determines which spatial relationships are captured
by the recurrent state.

Each direction $k$ defines a scan permutation $\pi_k$ --- a fixed
reordering of spatial positions that determines the traversal order.
Applying $\pi_k$ to $\mathbf{h}$ produces an independent token sequence
$\mathbf{s}_k = \pi_k(\mathbf{h}) \in \mathbb{R}^{N \times C}$, where
$N = H \cdot W$ is the number of spatial positions at the current
resolution stage. After the Mamba SSM processes $\mathbf{s}_k$, the
inverse permutation $\pi_k^{-1}$ restores spatial alignment. In the final experiments, $K = 3$ out of 6 possible scan
directions were used: a left-to-right row scan, a top-to-bottom column scan,
and an anti-diagonal scan. 
The effect of the number of scan directions $K$ is dataset- and
resolution-dependent: increasing $K$ does not yield consistent gains across
all settings, suggesting that the benefit of additional directions depends on
the geometric structure of the training data. This is examined in detail in
Section~\ref{sec:benchmark}.

\paragraph{Intra-block rotation.}
Before each DLR block, the feature map $\mathbf{h}$ is randomly rotated
by $180 ^\circ$. After the block's unscan, the rotation is inverted, restoring
the canonical spatial orientation. This is not a data augmentation
strategy: because the inversion is applied within the same forward pass,
the output spatial layout is unchanged. The purpose is to improve gradient
stability during training --- a $180 ^\circ$ rotation effectively reverses the
scan order of each direction, so the recurrent state receives gradients
from both forward and backward traversals, making the gradient signal more
consistent across training steps. No additional parameters are introduced.
The rotation mode is sampled uniformly from ${0^\circ, 180^\circ}$ during training.

Reversing the scan traversal order to expose the recurrent state to both
forward and backward gradient paths has been explored in related work:
LBMamba~\cite{lbmamba2025} and Adventurer~\cite{wang2025adventurer} reverse
the sequence order between consecutive layers to approximate bidirectional
scanning without additional SSM blocks.
Tang~et~al.~\cite{tang2024rotatetoscan} apply spatial rotation before
scanning in a medical segmentation context, motivated by coverage rather
than gradient flow. In contrast, our $180^\circ$ rotation is applied and
inverted within a single forward pass, leaving the output layout unchanged
--- its sole purpose is to expose the SSM recurrence to reversed gradient
paths during backpropagation.

Table~\ref{tab:ablation_rot} quantifies the effect of rotation on both
generation quality and gradient flow, evaluated on AFHQ $128{\times}128$.
Rotation yields a consistent improvement in FID and recall, a pattern that
holds across resolutions and datasets. Gradient flow analysis over
epochs~$0$--$200$ reveals that rotation increases mean gradient norms in
the deeper Mamba stages: the ratio between the two variants reaches
$1.23\times$ at the $64{\times}64$ stage, while shallower stages remain
largely unaffected. The coeficient 
of variation (CV $= \sigma/\mu$) measures the relative variability of 
gradient norms across training steps. CV remains comparable between 
the two variants across all stages, confirming that rotation does not 
destabilise gradient dynamics.

\begin{table}[h]
\centering
\small
\caption{Ablation: effect of $180^\circ$ rotation on AFHQ $128{\times}128$
(top) and mean gradient norm per Mamba block averaged over
all epochs (bottom).}
\label{tab:ablation_rot}
\begin{subtable}{\linewidth}
\centering
\resizebox{\linewidth}{!}{%
\begin{tabular}{l cccccc | cccccc}
\toprule
& \multicolumn{6}{c|}{Global} & \multicolumn{6}{c}{Mean Class} \\
\cmidrule(lr){2-7}\cmidrule(lr){8-13}
Rotation & FID $\downarrow$ & KID $\downarrow$ & P $\uparrow$ & R $\uparrow$ & D $\uparrow$ & C $\uparrow$
         & FID $\downarrow$ & KID $\downarrow$ & P $\uparrow$ & R $\uparrow$ & D $\uparrow$ & C $\uparrow$ \\
\midrule
None
  & 16.58 & 7.09 & .567 & .364 & 0.360 & .443
  & 18.46 & 11.70 & .543 & .318 & 0.327 & .418 \\
$\{0^\circ, 180^\circ\}$ (ours)
  & \textbf{11.66} & \textbf{2.71} & \textbf{.822} & \textbf{.407} & \textbf{1.144} & \textbf{.748}
  & \textbf{12.93} & \textbf{5.67} & \textbf{.826} & \textbf{.394} & \textbf{1.125} & \textbf{.744} \\
\bottomrule
\end{tabular}}
\end{subtable}

\vspace{0.5em}

\begin{tabular}{l cc cc c}
\toprule
& \multicolumn{2}{c}{None} & \multicolumn{2}{c}{$\{0^\circ, 180^\circ\}$ (ours)} & \\
\cmidrule(lr){2-3}\cmidrule(lr){4-5}
Mamba Block & mean norm & CV & mean norm & CV & ratio $\uparrow$ \\
\midrule
Tokenizer  & 0.02947 & 0.121 & 0.02713 & 0.118 & 0.921$\times$ \\
$8{\times}8$   & 0.01911 & 0.156 & 0.01862 & 0.205 & 0.974$\times$ \\
$16{\times}16$ & 0.00876 & 0.231 & 0.00955 & 0.254 & 1.091$\times$ \\
$32{\times}32$ & 0.00346 & 0.239 & 0.00427 & 0.230 & 1.233$\times$ \\
$64{\times}64$ & 0.00130 & 0.193 & 0.00160 & 0.215 & 1.232$\times$ \\
\bottomrule
\end{tabular}
\end{table}

\subsubsection{Class and Latent Conditioning per Direction}
\label{sec:conditioning}

\paragraph{Latent decomposition.}
The directional routing vector $\mathbf{z}_\text{dir} \in \mathbb{R}^{K
\cdot D_\text{dir}}$ is partitioned into $K$ equal subvectors
$\mathbf{z}_\text{dir} = [\mathbf{z}_\text{dir}^1 \mid \cdots \mid
\mathbf{z}_\text{dir}^K]$, where $\mathbf{z}_\text{dir}^k \in
\mathbb{R}^{D_\text{dir}}$. Each subvector $\mathbf{z}_\text{dir}^k$ is
exclusively associated with scan direction $k$ and does not influence any
other direction.

\paragraph{Per-direction class embedding.}
Each scan direction $k$ maintains a dedicated class embedding table
$E_k : \{1, \dots, C\} \to \mathbb{R}^{D_e}$. Given class label $y$,
the direction-specific embedding is $\mathbf{e}_k = E_k(y)$. Using
separate embedding tables per direction is a deliberate choice: different
scan axes capture spatially distinct features, and the class signal
relevant to each axis may differ. The DLR conditioning input for
direction $k$ is formed by concatenating the latent subvector with its
class embedding: $\mathbf{u}_k = [\mathbf{z}_\text{dir}^k \;\|\;
\mathbf{e}_k] \in \mathbb{R}^{D_\text{dir} + D_e}$.

\paragraph{Affine modulation.}
The joint conditioning vector $\mathbf{u}_k$ captures both the
direction-specific latent structure and the class identity. To translate
this into a conditioning signal for the Mamba SSM, a direction-specific
MLP projects $\mathbf{u}_k$ to a pair of feature-wise affine parameters
$[\tilde{\boldsymbol{\gamma}}_k,\, \tilde{\boldsymbol{\beta}}_k] =
\mathrm{MLP}_k(\mathbf{u}_k)$, where
$\tilde{\boldsymbol{\gamma}}_k,\, \tilde{\boldsymbol{\beta}}_k \in
\mathbb{R}^{C}$. This is the core of the DLR mechanism: rather than
injecting the conditioning signal additively into the feature map (as in
standard FiLM~\cite{perez2018film}), the modulation is applied to the
\emph{token sequence} $\mathbf{s}_k$ before it enters the SSM recurrence.
This means the class and latent information directly shapes what the
recurrent state integrates along direction $k$, rather than post-hoc
shifting the output. Each direction receives an independent
$\mathrm{MLP}_k$, so the conditioning function can specialize to the
spatial statistics of that scan axis.

To improve training stability, \emph{residual routing} is applied: the
raw MLP output is passed through a clipped tanh and offset so that the
modulation starts close to identity:
\begin{equation}
  \boldsymbol{\gamma}_k = \tanh(\tilde{\boldsymbol{\gamma}}_k)
    \cdot \delta_\ell + 1, \qquad
  \boldsymbol{\beta}_k  = \tanh(\tilde{\boldsymbol{\beta}}_k)
    \cdot \delta_\ell,
  \label{eq:residual_routing}
\end{equation}
where $\delta_\ell > 0$ is a resolution-dependent clip value. The $+1$ offset ensures
$\boldsymbol{\gamma}_k \approx 1$ at initialisation, so the DLR block
acts as identity early in training and the modulation is learned
progressively. The conditioned token sequence fed into the SSM is then:
\begin{equation}
  \hat{\mathbf{s}}_k = \boldsymbol{\gamma}_k \odot \mathbf{s}_k
  + \boldsymbol{\beta}_k,
  \label{eq:modulation}
\end{equation}
where $\boldsymbol{\gamma}_k$ and $\boldsymbol{\beta}_k$ are broadcast
over the sequence length dimension $N$.

\begin{figure}[h]
  \centering
  \includegraphics[width=0.8\linewidth]{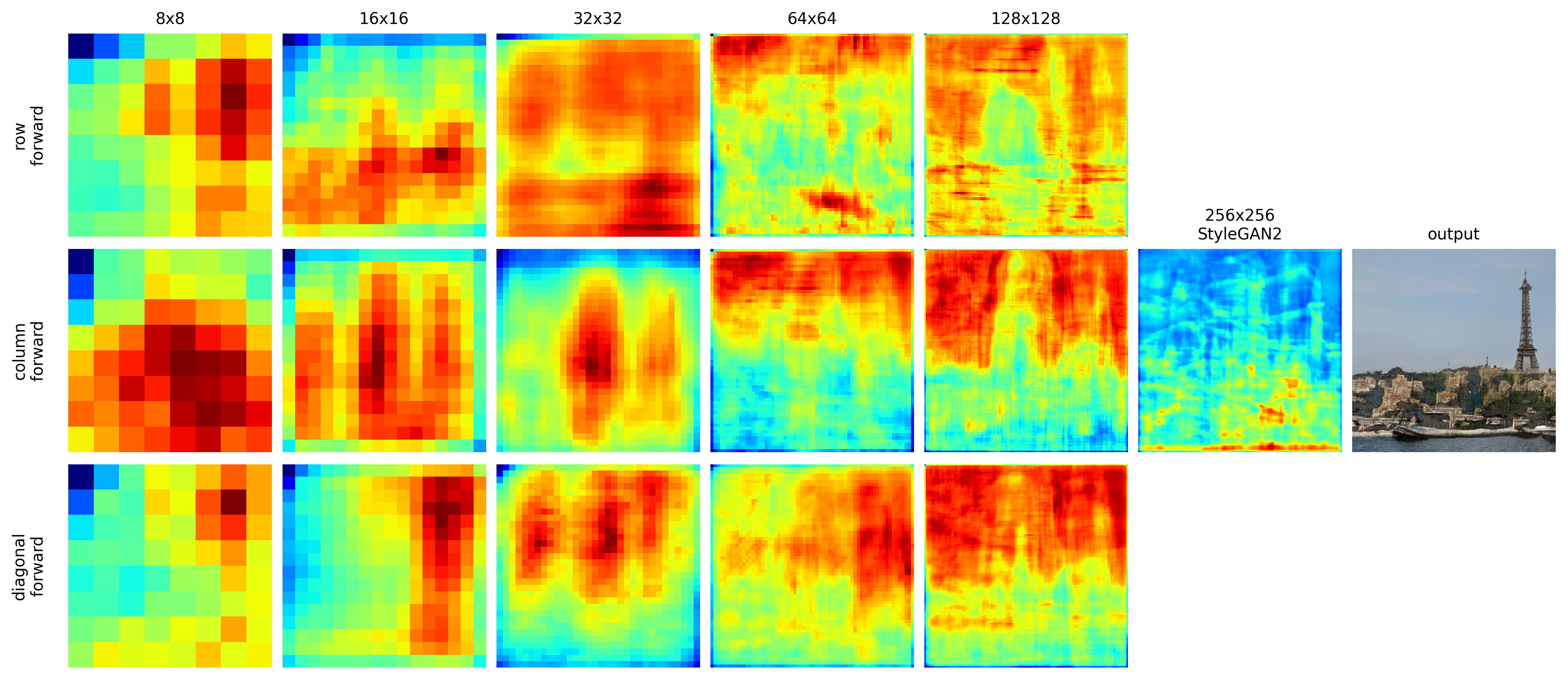}
  \caption{Per-direction feature maps (mean absolute activation, averaged
  over 10 channels) at each resolution stage, for a representative
  sample from the LSUN \emph{tower} class (from \textit{bridge, church, tower classes}) in Mamba blocks. Each row corresponds to one
  scan direction; each column to one resolution stage. At low resolutions
  directions capture complementary aspects of global structure; at higher
  resolutions the directional geometry becomes explicit, with row, column,
  and diagonal activations forming spatially oriented patterns consistent
  with their respective scan axes.}
  \label{fig:feature_maps}
\end{figure}

Figure~\ref{fig:feature_maps} visualises the mean absolute activations
per scan direction at each resolution stage, confirming that independent
$\mathrm{MLP}_k$ conditioning leads to direction-specific specialisation.
At low resolutions ($8{\times}8$, $16{\times}16$), directions capture
complementary aspects of global structure --- the column scan concentrates
on the vertical extent of the subject, the row scan on horizontal layout.
At higher resolutions ($64{\times}64$, $128{\times}128$) the directional
geometry becomes explicit: row activations form horizontal bands, column
activations vertical bands, and diagonal activations oriented texture.

\subsubsection{Direction Weighting}
\label{sec:weighting}

After each directional Mamba block processes its modulated sequence, the
outputs are combined via a learned weighted sum $\mathbf{w} \in
\mathbb{R}^K$. To compute these weights, a small routing network maps the
directional features $\mathbf{z}_\text{dir}$ to per-direction logits using
a weight matrix $\mathbf{W}_z$. These logits are added to learnable base
weights $\mathbf{b} \in \mathbb{R}^K$ and a shared class embedding
$\mathbf{e}_y = \mathrm{Embed}(y) \in \mathbb{R}^{D_e}$ projected by a
weight matrix $\mathbf{W}_y$. Both $\mathbf{W}_z$ and $\mathbf{W}_y$ are
initialized to zero so that routing starts uniform. Finally, the total sum
is scaled by a temperature parameter $\tau$ and normalized via the softmax
function, denoted as $\sigma_S$:

\begin{equation}
  \mathbf{w} = \sigma_S((\mathbf{b} + \mathbf{W}_z\mathbf{z}_\text{dir} +
  \mathbf{W}_y\mathbf{e}_y) / \tau).
  \label{eq:softmax}
\end{equation}

The class label $y$ enters the direction weighting jointly with
$\mathbf{z}_\text{dir}$. This is a deliberate design choice: the class
signal serves a dual role --- it modulates \emph{what is generated} along
each direction (via $\boldsymbol{\gamma}_k$, $\boldsymbol{\beta}_k$ in
Eq.~\ref{eq:modulation}) and simultaneously steers \emph{how much each
direction contributes} to the output. Class-conditioned weighting allows
the model to learn that certain scan axes are more informative for specific
categories --- for instance, a class with strong vertical structure may
consistently up-weight the column scan. Intra-class diversity is preserved
through $\mathbf{z}_\text{dir}$, which provides a sample-specific
perturbation of the class-determined base routing. The degenerate case in which all directions are set identically --- eliminating the geometric basis for specialisation --- is examined empirically in Appendix ~\ref{app:direction_diversity}.

Figure~\ref{fig:routing_evolution} tracks the evolution of per-direction
routing weights across resolution stages over the course of training
(AFHQ $256{\times}256$, 3-direction model). Weights are initialised close
to uniform ($1/K = 0.33$, dashed red line) and diverge as training
progresses, confirming that the direction weighting network learns
non-trivial, resolution-dependent routing. The column scan acquires the
highest weight at $16{\times}16$ (0.54), suggesting that
vertically-oriented structure is most informative at low resolutions. The
row scan is suppressed at $16{\times}16$ but recovers at higher
resolutions, while the diagonal scan peaks at $32{\times}32$ before
declining. The most pronounced specialisation occurs at the lowest
resolution stages, where the primary class-specific structure is
established. This is consistent with the sensitivity analysis in
Table~\ref{tab:lpips_transposed}, which shows that class variations
produce the highest perceptual impact at $8{\times}8$ (LPIPS $= 0.612$)
and $16{\times}16$ (LPIPS $= 0.464$), with substantially lower sensitivity
at higher resolutions. LPIPS~\cite{zhang2018lpips} measures perceptual
distance between generated images under isolated component variations ---
higher values indicate that changing a given factor produces more visually
distinct outputs, making it a direct proxy for the influence of each
component on the generated result. The direction weighting network thus
concentrates its class-aware routing signal precisely where it has the
greatest influence on the generated output.

\begin{figure}[h]
  \centering
  \includegraphics[width=\linewidth]{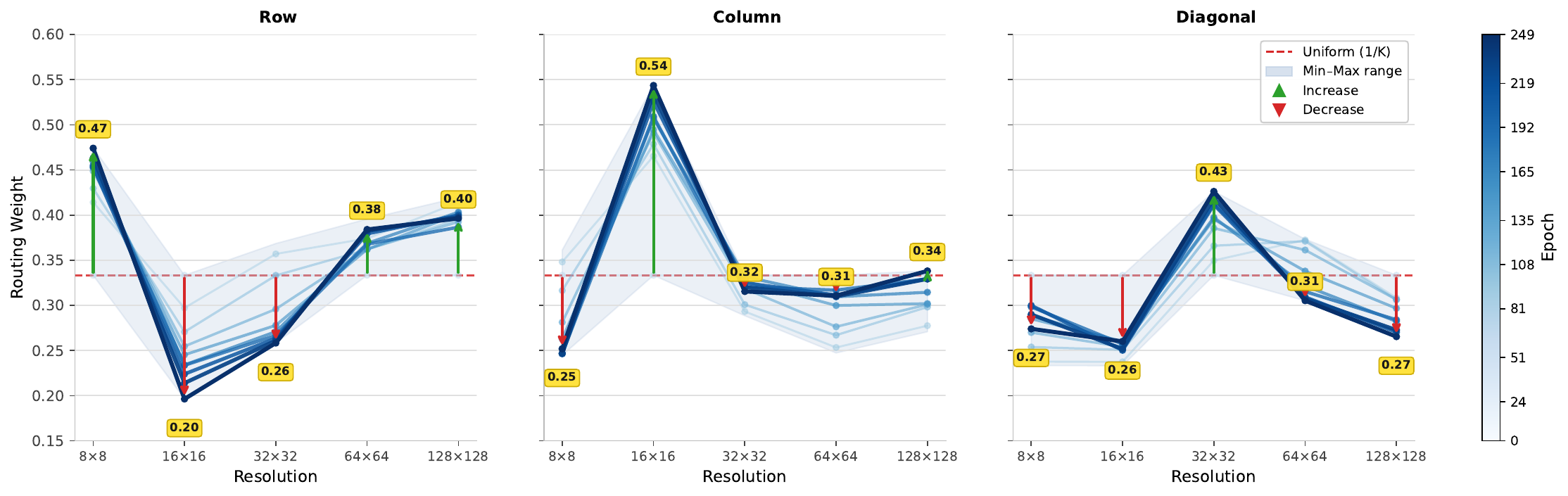}
  \caption{Evolution of per-direction routing weights across resolution
  stages during training (AFHQ $256{\times}256$, 3-direction model).
  Each line corresponds to a training snapshot; colour indicates epoch
  (light = early, dark = late). The dashed red line marks the uniform
  weight $1/K$. Arrows indicate the direction of change between the
  first and last snapshot. Weights diverge from uniform and specialise
  by resolution, with the column scan dominating at $16{\times}16$ and
  the diagonal scan peaking at $32{\times}32$.}
  \label{fig:routing_evolution}
\end{figure}
\subsubsection{Mamba Blocks and Output Aggregation}
\label{sec:aggregation}

Table~\ref{tab:lpips_transposed} reports the perceptual impact (LPIPS) of independently perturbing the class and latent representations in each model component. In the class experiment, the learned class is fully substituted with that of a target class in a selected subset of components - a single scan direction, a single resolution stage, or the tokenizer and CNN refinement block - while all remaining components, $y$, and $\mathbf{z}$ are kept fixed. In the latent experiment, the corresponding fragment of $\mathbf{z}$ is replaced with an independent Gaussian sample in the same subset of components, while $y$ and the remaining parts of $\mathbf{z}$ are kept fixed. In both cases, LPIPS is computed between the unmodified output and the output after the perturbation, and averaged over 10 randomly sampled latent vectors. The class experiment
additionally averages over all six ordered class pairs.
A higher LPIPS value indicates a higher perceptual difference in the generated image, and thus a stronger influence of the perturbed component on the output.

\begin{table}[H]
\centering
\small
\setlength{\tabcolsep}{4pt}
\caption{Sensitivity analysis: Perceptual impact (LPIPS $\uparrow$) of Class vs.\ Latent
  variations across model components.}
\label{tab:lpips_transposed}
\begin{tabular}{l ccc ccccc ccc}
\toprule
 & \multicolumn{3}{c}{General} & \multicolumn{5}{c}{Mamba Stages} & \multicolumn{3}{c}{Mamba Directions} \\
\cmidrule(lr){2-4}\cmidrule(lr){5-9}\cmidrule(lr){10-12}
Factor & All dir & Base & CNN & $8{\times}8$ & $16{\times}16$ & $32{\times}32$ & $64{\times}64$ & $128{\times}128$ & Col & Row & Diag \\
\midrule
Class  & 0.676 & 0.044 & 0.047 & 0.612 & 0.464 & 0.190 & 0.149 & 0.144 & 0.484 & 0.455 & 0.485 \\
Latent & 0.682 & 0.621 & ---   & 0.628 & 0.496 & 0.091 & 0.322 & 0.290 & 0.608 & 0.591 & 0.573 \\
\bottomrule
\end{tabular}
\end{table}

\begin{figure}[h!]
  \centering
  \includegraphics[width=0.48\linewidth]{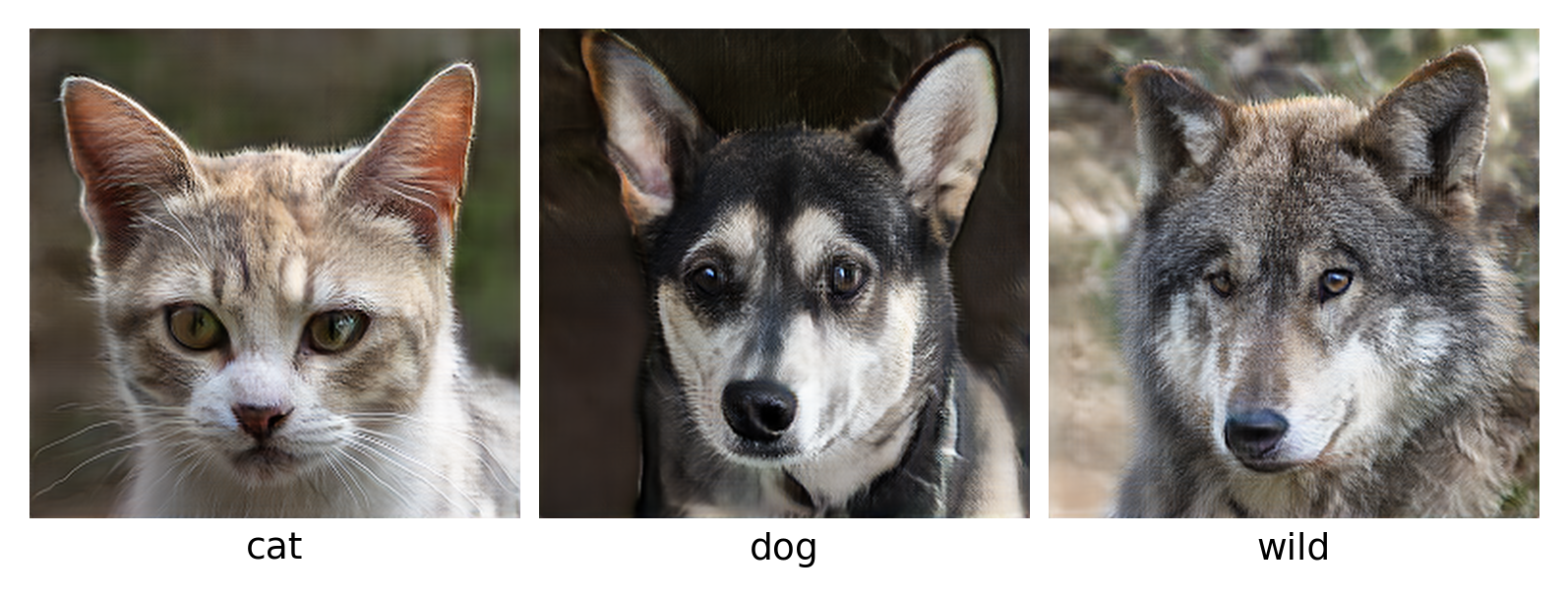}\hfill
  \includegraphics[width=0.48\linewidth]{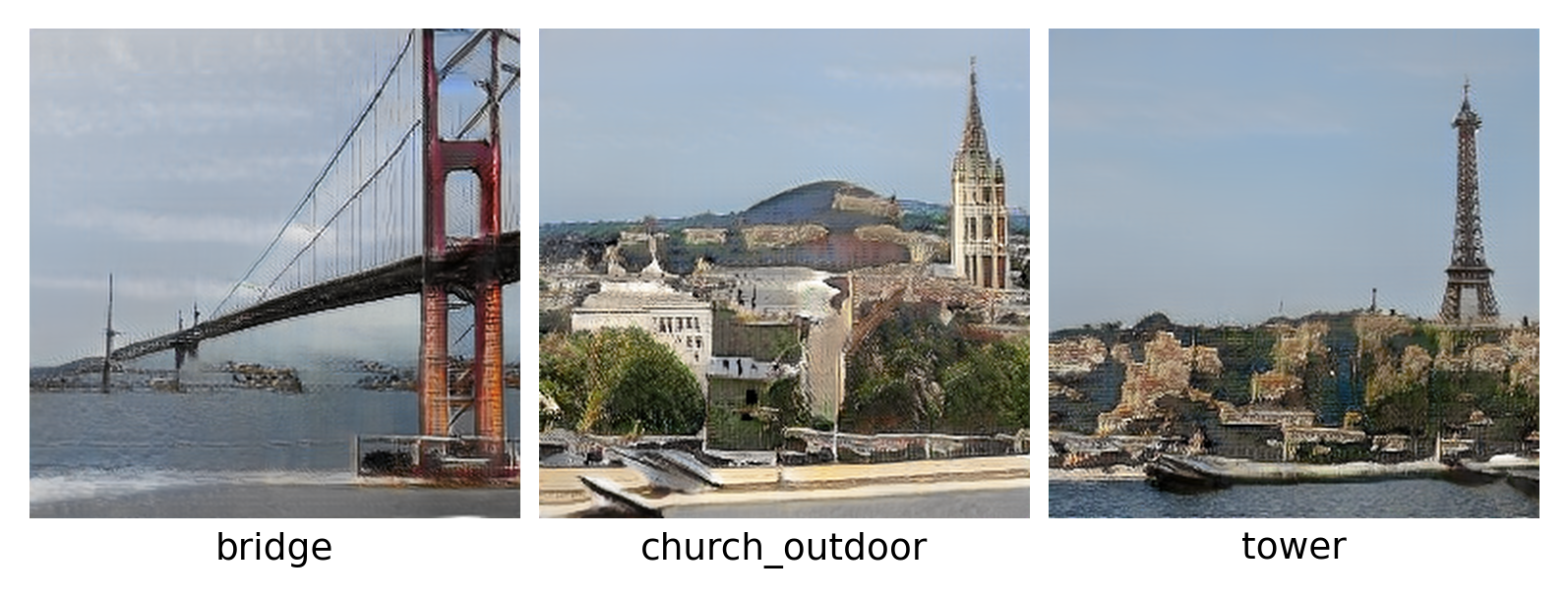}
  \caption{The same latent vector $\mathbf{z}$ generated with three different
  class labels. The spatial layout is preserved across classes, confirming that
  $\mathbf{z}_\text{base}$ controls composition independently of class identity. Left: AFHQ $256 \times 256$ , right : LSUN $256 \times 256 $.}
  \label{fig:classswap}
\end{figure}

\paragraph{Disentanglement of the latent space.}
The two components of $\mathbf{z}$ serve complementary roles in the
generated image, each governing a distinct aspect of the synthesis process. $\mathbf{Z}_\text{base}$ controls global composition and structure through
the tokenizer, while $\mathbf{z}_\text{dir}$ shapes what the SSM integrates
along each scan direction through the affine modulations
$(\boldsymbol{\gamma}_k, \boldsymbol{\beta}_k)$. The class label $y$ can be
swapped independently of $\mathbf{z}$, producing a semantic change while
preserving the spatial layout --- as illustrated in Figure~\ref{fig:classswap}.

\paragraph{Class override experiments.}

\begin{figure*}[h!]
  \centering
  \includegraphics[width=\linewidth]{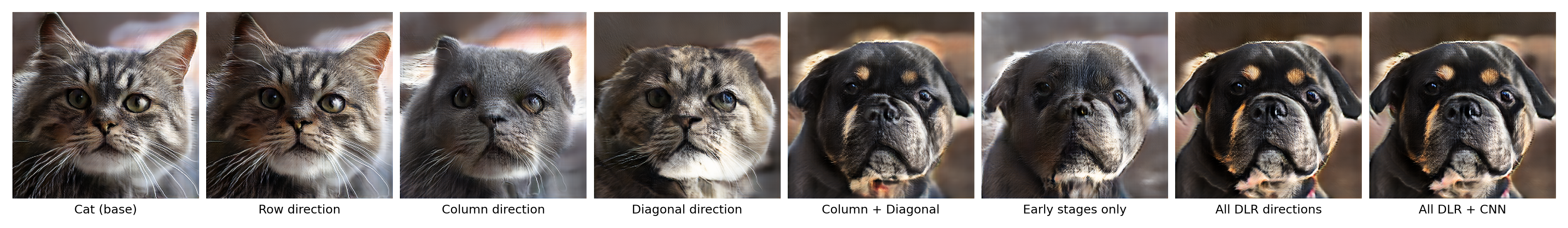}

  \vspace{1em}
  \includegraphics[width=\linewidth]{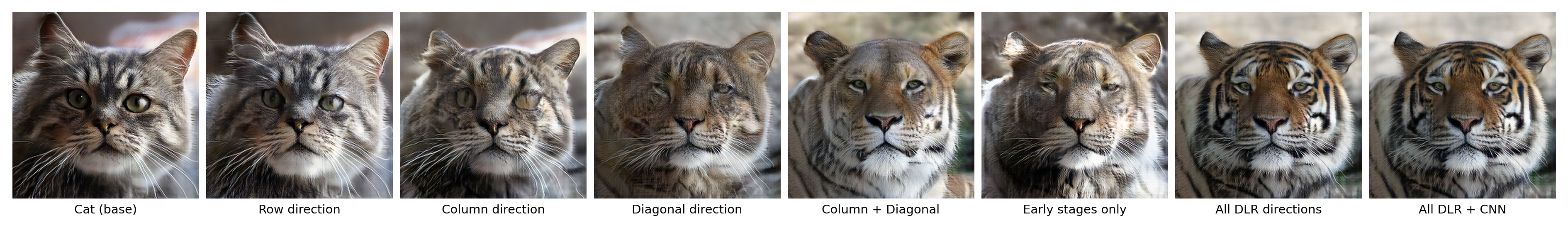}
  \caption{Partial class override experiments on AFHQ $256 \times 256$. \textbf{Top:} base class
  \emph{cat}, target class \emph{dog}. \textbf{Bottom:} base class \emph{cat},
  target class \emph{wild}. Each column replaces a different subset of
  per-direction classes with the target class, keeping $\mathbf{z}$ fixed.
  The row direction shows little influence on class conversion; the majority
  of class information is modulated by the blocks at low resolutions.}
  \label{fig:class_override}
\end{figure*}

To investigate how individual scan directions carry class information, we
conduct experiments in which only a selected subset of per-direction class
embeddings is replaced with those of a target class, while the remaining
directions and $\mathbf{z_{base}}$ are kept fixed. Figure~\ref{fig:class_override}
shows results for the pairs \emph{cat}$\to$\emph{dog} and
\emph{cat}$\to$\emph{wild} --- overriding a single direction produces a
gradual semantic transformation, and only replacing all directions yields a
complete class conversion. The perceptual distance LPIPS between the base and
overridden image, averaged over $N$ random latent vectors and six class pairs
from \{cat, dog, wild\}, is summarised in Table~\ref{tab:lpips_transposed}.

\paragraph{Directional latent swap.}

\begin{figure*}[h!]
  \centering
  \includegraphics[width=\linewidth]{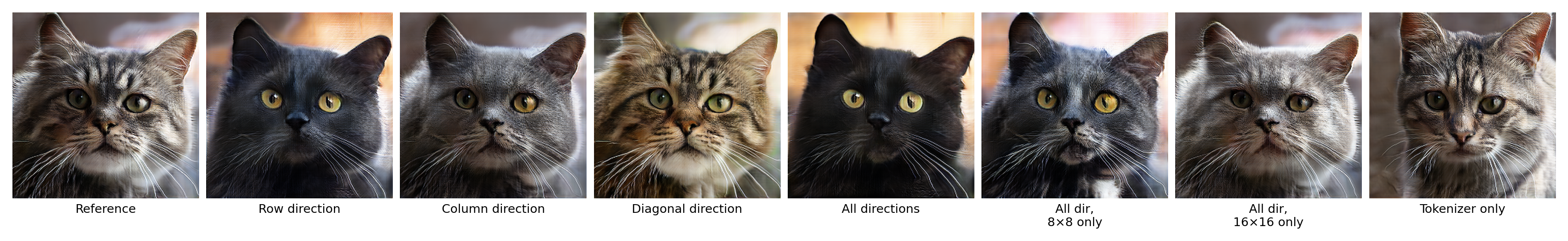}
  \caption{Swapping directional subvectors $\mathbf{z}_\text{dir}^k$ between
  samples. Each column replaces the subvector of one direction while keeping
  all other components of $\mathbf{z}$ fixed.}
  \label{fig:latent_swap}
\end{figure*}

Figure~\ref{fig:latent_swap} shows the effect of randomly swapping the
subvector $\mathbf{z}_\text{dir}^k$ between samples while keeping all other
components fixed. Replacing the subvector of a single direction produces a
localised change in style or texture characteristic of that scan axis, while
the global structure of the image remains unchanged --- consistent with the
low LPIPS sensitivity of higher-resolution stages to latent variations seen
in Table~\ref{tab:lpips_transposed}.

\begin{figure}[htbp]
  \centering
  \includegraphics[width=0.48\linewidth]{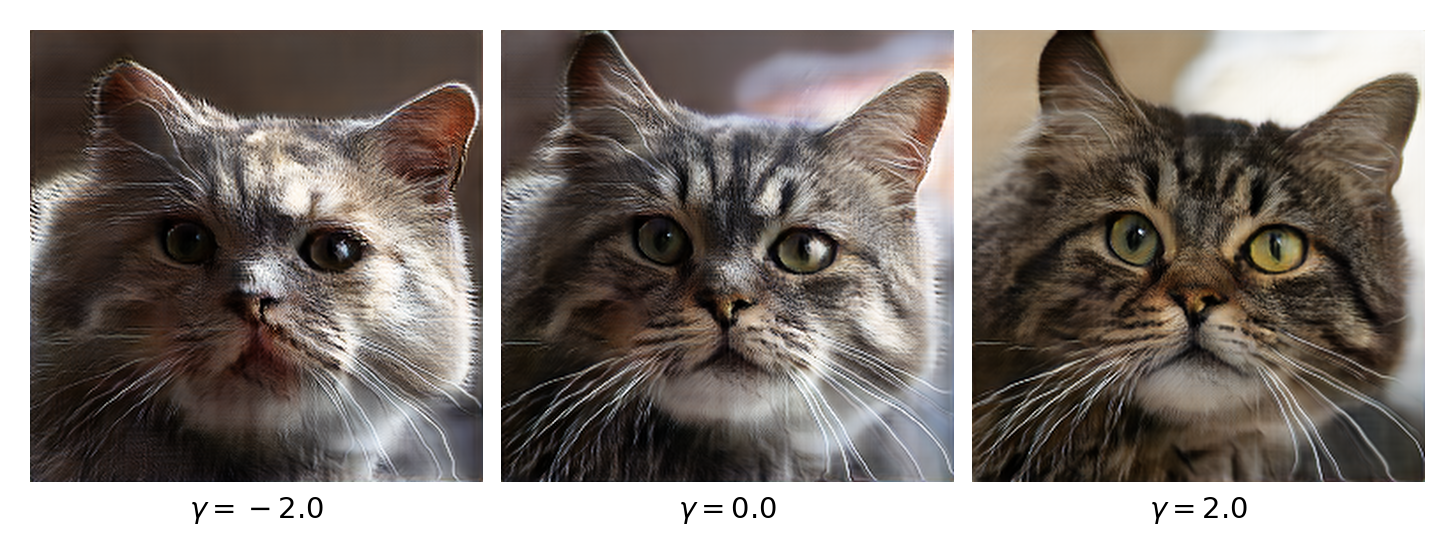}\hfill
  \includegraphics[width=0.48\linewidth]{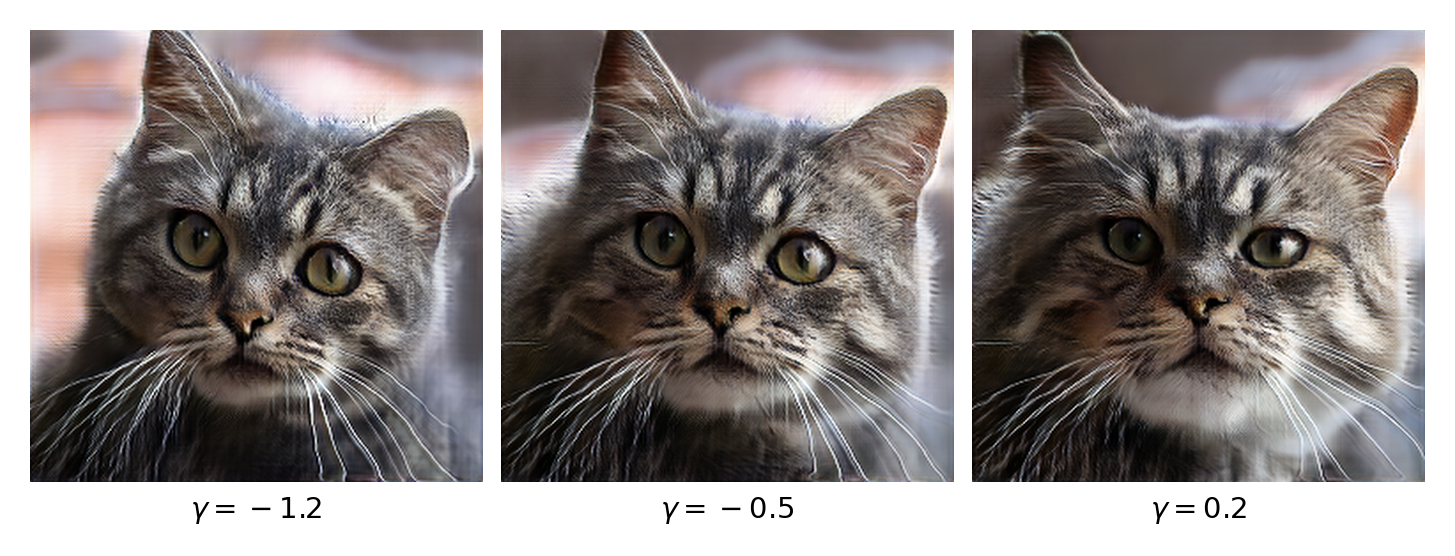}
  \caption{Effect of noise perturbation on latent space components.
  \textbf{Left:} perturbation of the column direction
  ($\mathbf{z}_\mathrm{dir}^\mathrm{col}$). \textbf{Right:} perturbation of
  the tokenizer component ($\mathbf{z}_\mathrm{base}$).}
  \label{fig:noise_perturbation}
\end{figure}

\paragraph{Noise perturbation.}
Figure~\ref{fig:noise_perturbation} shows the effect of Gaussian perturbation
applied to $\mathbf{z}_\text{dir}^\text{col}$ and $\mathbf{z}_\text{base}$.
Perturbing $\mathbf{z}_\text{base}$ alters the spatial layout and overall
scene composition, as expected from its high LPIPS sensitivity
(Table~\ref{tab:lpips_transposed}). Directional components learn finer
details: subtle geometry, colouring, and surface structure. In particular,
perturbing $\mathbf{z}_\text{dir}^\text{col}$ gradually modifies texture and
sharpness while preserving subject identity.

Corresponding results for all above experiments for the LSUN outdoor dataset are provided in Appendix~\ref{app:lsun_conditioning}.

\section{Results}
\label{sec:benchmark}

This section presents the results of DSS-GAN across five dataset configurations at varying
resolutions and scan direction settings. As a reference point, we use
StyleGAN2-ADA~\cite{karras2020ada}, a model combining the StyleGAN2 generator with adaptive
discriminator augmentation, which established state-of-the-art results on FFHQ and AFHQ at the
time of its publication~\cite{karras2020ada}. Since the discriminator is identical in both models,
any difference in performance can be attributed directly to the generator. We evaluate on four
datasets: AFHQ~\cite{choi2020stargan} (3 classes: cat, dog, wild, approximately 5k samples
per class),
FFHQ~\cite{karras2019stylegan} (unconditional, 70k samples), LSUN~\cite{yu2015lsun} evaluated on two
distinct subsets --- indoor scenes (LSUN Rooms: bedroom, kitchen; 15k samples per class) at $128{\times}128$ and outdoor
architectural scenes (bridge, church, tower; 15k samples per class) at $256{\times}256$ --- and
CelebA~\cite{liu2015celeba} (10 attribute classes, 10k samples per class). All subsets use a
fixed sample selection across all experiments. All models were trained
from scratch.

StyleGAN2-ADA training configurations define a set of hyperparameters --- network depth and
width, $R_1$ regularisation strength, learning rate schedule --- tuned to a specific resolution and
dataset~\cite{karras2020adagithub}. For FFHQ we used \texttt{paper256} that is dedicated to this dataset on $256 \times256$ resolution. For AFHQ, LSUN, and CelebA we compared \texttt{auto}, \texttt{paper256}, and \texttt{paper512}, selecting the
configuration with the lowest FID --- in all cases this was \texttt{paper512}. This configuration is dedicated to AFHQ dataset on $512 \times 512$ resolution. DSS-GAN was trained
with identical hyperparameters across all datasets at a given resolution; as resolution increases,
$R_1$ regularisation strength and latent space dimensionality are increased, while the number of
feature map channels is reduced. Both models were trained with a batch size of 96 at
$128{\times}128$ and $256{\times}256$, and 48 at $512{\times}512$, with ADA augmentation
disabled, eliminating augmentation as a confounding variable. ADA could be applied to DSS-GAN
as well but was deliberately excluded. 

Sample quality was assessed using FID~\cite{heusel2017fid} and KID~\cite{binkowski2018kid} ($\times 10^3$), which measure distributional similarity between generated and real samples. We additionally report Precision~(P), Recall~(R), Density~(D), and Coverage~(C)
following~\cite{kynkaanniemi2019precision,naeem2020dc}, which decompose sample quality into
fidelity~(P,\,D) and diversity~(R,\,C) components. All metrics are computed over 50k samples. Density and Coverage extend Precision and Recall by replacing binary membership with a count-based estimator (Density) and a coverage indicator (Coverage), yielding metrics that are more robust to outliers in the feature space. Both models were evaluated with the same script and the same set of
real reference images. FID and KID values are consistent with those produced by the original
StyleGAN2-ADA evaluation script~\cite{karras2020adagithub}, with a maximum deviation of 0.3
FID points and 0.2 KID points across best checkpoints.

\paragraph{Results.}
Results are reported in order of increasing resolution. Per-class metrics are provided for AFHQ
and LSUN; CelebA results are averaged over the 10 attribute classes; FFHQ is reported as a
single unconditional result. KID values are reported multiplied by $10^3$ throughout. Where not
stated otherwise, the 1-dir variant uses a left-to-right row scan.

\begin{table}[ht]
\centering
\small
\caption{Results on LSUN Rooms $128{\times}128$ (classes: Bedroom, Kitchen).}
\label{tab:lsun_rooms}
\resizebox{\textwidth}{!}{%
\begin{tabular}{lcccccc|cccccc|cccccc}
\toprule
& \multicolumn{6}{c|}{Global} & \multicolumn{6}{c|}{Bedroom} & \multicolumn{6}{c}{Kitchen} \\
Model & FID$\downarrow$ & KID$\downarrow$ & P$\uparrow$ & R$\uparrow$ & D$\uparrow$ & C$\uparrow$
      & FID$\downarrow$ & KID$\downarrow$ & P$\uparrow$ & R$\uparrow$ & D$\uparrow$ & C$\uparrow$
      & FID$\downarrow$ & KID$\downarrow$ & P$\uparrow$ & R$\uparrow$ & D$\uparrow$ & C$\uparrow$ \\
\midrule
StyleGAN2-ADA
& \underline{20.31} & \underline{10.10} & \underline{.59} & .25 & \textbf{.94} & .65
& 26.06 & 11.51 & \underline{.58} & .24 & \textbf{.85} & .61
& 26.64 & \underline{12.70} & \underline{.62} & .27 & \textbf{.97} & .58 \\
DSS-GAN 3-dir
& 23.22 & 13.23 & .52 & .28 & .62 & .67
& 25.16 & 13.45 & .52 & .28 & .62 & .67
& 31.29 & 20.28 & .52 & .28 & .62 & .67 \\
DSS-GAN 2-dir (row + column)
& \textbf{17.79} & \textbf{8.79} & \textbf{.64} & .20 & \underline{.88} & \textbf{.75}
& \underline{21.56} & \textbf{9.61} & \textbf{.59} & .26 & \underline{.78} & \underline{.76}
& \textbf{23.32} & \textbf{12.30} & \textbf{.71} & .18 & \textbf{.97} & \textbf{.78} \\
DSS-GAN 1-dir (row)
& \underline{20.28} & 11.93 & .55 & \underline{.30} & .66 & \underline{.73}
& \textbf{21.39} & \underline{10.06} & .55 & \underline{.35} & .69 & \textbf{.79}
& 28.78 & 19.35 & .59 & \underline{.27} & .62 & .72 \\
DSS-GAN 1-dir (column)
& 20.30 & 10.73 & .58 & .23 & .65 & .68
& 24.97 & 11.95 & \underline{.58} & .24 & .69 & .69
& \underline{25.83} & 14.43 & .60 & .18 & .70 & \underline{.75} \\
DSS-GAN 1-dir (diagonal)
& 22.74 & 14.44 & .50 & \textbf{.31} & .56 & .67
& 23.80 & 12.32 & .47 & \textbf{.40} & .47 & .69
& 32.76 & 22.81 & .54 & \textbf{.28} & .55 & .68 \\
\bottomrule
\end{tabular}%
}
\end{table}

\paragraph{LSUN Rooms $128{\times}128$.}
The LSUN Rooms dataset~\cite{yu2015lsun} contains two classes of indoor scenes (bedroom,
kitchen), characterised by rectangular geometry dominated by two principal spatial directions.
Table~\ref{tab:lsun_rooms} shows that the 2-dir variant achieves the lowest FID and KID among
all DSS-GAN variants and outperforms StyleGAN2-ADA globally. The 3-dir variant underperforms
relative to 2-dir despite the larger number of scan directions --- the diagonal direction does not
correspond to the dominant geometric structure of the dataset and introduces a conditioning signal
misaligned with the data. Among single-direction variants (1-dir), the diagonal achieves the
weakest results across all metrics except Recall, directly confirming that scan direction
effectiveness is a property of the dataset rather than of the direction itself. Some DSS-GAN variants
show consistently higher Recall than StyleGAN2-ADA, indicating broader coverage of the real data
distribution.

\begin{table}[ht]
\centering
\scriptsize
\caption{Results on CelebA $128{\times}128$. Classes: Bald, Black Hair, Blond Hair, Eyeglasses,
Goatee, Male, Mustache, Smiling, Wearing Hat, Young (10k samples per class).}
\label{tab:celeba}
\begin{tabular}{lcccccc|cccccc}
\toprule
& \multicolumn{6}{c|}{Global} & \multicolumn{6}{c}{Mean class} \\
Model & FID$\downarrow$ & KID$\downarrow$ & P$\uparrow$ & R$\uparrow$ & D$\uparrow$ & C$\uparrow$
      & FID$\downarrow$ & KID$\downarrow$ & P$\uparrow$ & R$\uparrow$ & D$\uparrow$ & C$\uparrow$ \\
\midrule
StyleGAN2-ADA & 15.63 & 10.91 & .56 & .38 & .55 & .68 & 14.30 & 10.28 & .59 & .33 & .53 & .70 \\
DSS-GAN 1-dir & \textbf{12.42} & \textbf{6.17} & \underline{.69} & \textbf{.56} & \underline{.85} & \textbf{.83} & \textbf{11.05} & \textbf{7.39} & \underline{.76} & \textbf{.57} & \underline{.90} & \underline{.89} \\
DSS-GAN 3-dir & \underline{13.40} & \underline{7.43} & \textbf{.74} & \underline{.52} & \textbf{.88} & \textbf{.83} & \underline{12.19} & \underline{8.91} & \textbf{.79} & \underline{.54} & \textbf{.91} & \textbf{.92} \\
\bottomrule
\end{tabular}
\end{table}

\paragraph{CelebA $128{\times}128$.}
The CelebA dataset~\cite{liu2015celeba} comprises 10 facial attribute classes (10k samples per
class). Human faces exhibit a relatively isotropic local structure --- skin, hair, and facial feature
textures do not display a dominant spatial axis. As shown in Table~\ref{tab:celeba}, the 1-dir
variant achieves lower FID and KID than the 3-dir variant, both globally and per class. With
isotropic data, additional scan directions do not contribute complementary spatial information, and
a simpler latent routing translates to better coverage of the sample space. Both DSS-GAN
variants achieve lower FID and KID alongside higher Precision, Recall, and Coverage relative to
StyleGAN2-ADA. Density increases from 0.55 to 0.88--0.91, indicating that the Mamba backbone
concentrates generated samples more tightly around the real data distribution even in the absence
of strong directional structure.

\begin{table}[ht]
\centering
\small
\caption{Results on AFHQ $256{\times}256$.}
\label{tab:afhq}
\resizebox{\textwidth}{!}{%
\begin{tabular}{lcccccc|cccccc|cccccc|cccccc}
\toprule
& \multicolumn{6}{c|}{Global} & \multicolumn{6}{c|}{Cat} & \multicolumn{6}{c|}{Dog} & \multicolumn{6}{c}{Wild} \\
Model & FID$\downarrow$ & KID$\downarrow$ & P$\uparrow$ & R$\uparrow$ & D$\uparrow$ & C$\uparrow$
      & FID$\downarrow$ & KID$\downarrow$ & P$\uparrow$ & R$\uparrow$ & D$\uparrow$ & C$\uparrow$
      & FID$\downarrow$ & KID$\downarrow$ & P$\uparrow$ & R$\uparrow$ & D$\uparrow$ & C$\uparrow$
      & FID$\downarrow$ & KID$\downarrow$ & P$\uparrow$ & R$\uparrow$ & D$\uparrow$ & C$\uparrow$ \\
\midrule
StyleGAN2-ADA
& \underline{13.16} & 4.13 & .75 & .21 & .84 & .64
& 11.04 & 4.70 & .78 & .14 & 1.15 & .78
& 33.27 & 18.49 & .73 & .25 & .55 & .43
& 7.56 & \underline{1.96} & \underline{.77} & .12 & .82 & \underline{.64} \\
DSS-GAN 1-dir
& 13.58 & \underline{3.37} & .85 & \underline{.35} & 1.14 & \underline{.67}
& \underline{9.74} & \underline{4.46} & .88 & \textbf{.26} & 1.41 & \underline{.87}
& \underline{27.34} & \underline{14.60} & \textbf{.85} & \textbf{.55} & .70 & \textbf{.53}
& \textbf{6.07} & 1.96 & \underline{.77} & \textbf{.23} & \underline{1.02} & .62 \\
DSS-GAN 2-dir (row+col)
& 22.50 & 5.85 & .87 & .16 & 1.14 & .53
& 20.09 & 10.04 & \textbf{.95} & .08 & \underline{1.91} & .74
& 41.37 & 23.01 & \underline{.86} & .32 & \underline{.78} & .44
& 19.61 & 7.94 & .73 & .06 & .66 & .38 \\
DSS-GAN 2-dir (row+diag)
& 22.76 & 5.17 & \textbf{.89} & .17 & \textbf{1.33} & .53
& 20.88 & 9.35 & \underline{.94} & .10 & \textbf{1.97} & .72
& 40.03 & 19.03 & .91 & .25 & \textbf{.90} & .45
& 21.90 & 8.67 & \underline{.77} & \underline{.13} & \textbf{1.02} & .41 \\
DSS-GAN 3-dir
& \textbf{10.29} & \textbf{2.39} & \underline{.88} & \textbf{.36} & \textbf{1.33} & \textbf{.74}
& \textbf{8.87} & \textbf{2.65} & .91 & \underline{.18} & 1.87 & \textbf{.89}
& \textbf{26.51} & \textbf{13.05} & .83 & \underline{.46} & .73 & \underline{.51}
& \underline{6.73} & \textbf{1.51} & \textbf{.87} & .09 & \textbf{1.74} & \textbf{.77} \\
\bottomrule
\end{tabular}%
}
\end{table}

\paragraph{AFHQ $256{\times}256$.}
The AFHQ dataset~\cite{choi2020stargan} contains three animal classes. As reported in Table~\ref{tab:afhq}, the 3-dir variant achieves FID 10.29 against 13.16
for StyleGAN2-ADA. The most pronounced improvement is in Density: 1.33 against 0.84 globally,
and 1.87 against 1.15 for the cat class --- indicating that generated samples not only lie closer to
the real data manifold but also cover it more uniformly. Fur texture, body contours, and coat
markings exhibit strong spatial correlations along multiple axes, providing a natural setting for the
DLR mechanism where each scan direction contributes complementary structural information.
Notably, both 2-dir variants (row+col and row+diag) underperform not only 3-dir but also
1-dir, with global FID degrading to 22.50 and 22.76 respectively, and Recall collapse for the cat class. Crucially, the two 2-dir variants perform comparably
despite their different geometric relationship: row+col provides orthogonal axes while
row+diag shares a horizontal component with the row direction, yet neither combination
recovers the quality of 1-dir. This suggests that the degradation does not stem from an
inappropriate choice of directions but from a structural property of $K{=}2$. One possible
interpretation is a routing stability failure: at $K{=}1$, DLR reduces to global conditioning
with no conflict between directions, whereas at $K{=}2$, two directions must share the latent
space and compete for routing weight --- without a third direction to mediate, the
decomposition may oscillate between inconsistent solutions, which could account for the
Recall collapse on the cat class. Appendix~\ref{app:direction_diversity} provides direct empirical evidence for this failure mode: training with $K{=}3$ identical row scans shows that routing weight specialisation
does not emerge, the class routing scale $|\alpha|$ remains near zero throughout training,
and the model collapses after epoch~108. At $K{=}3$, routing stabilises because three directions
form a sufficiently rich basis for the isotropic geometry of AFHQ, allowing each direction
to specialise in a distinct aspect of the spatial structure.

This result contrasts with LSUN Rooms, where 2-dir is optimal: the rectangular geometry
of indoor scenes admits a natural two-axis decomposition aligned with the horizontal and
vertical scan directions, requiring no third direction to stabilise routing. The optimal
number of scan directions is thus a property of the dataset geometry rather than a
hyperparameter to maximise. The cat class exhibits consistently low Recall across all
DSS-GAN variants despite high Precision and Density, suggesting that the generator covers
the real distribution with high fidelity but does not span its full extent --- a pattern
likely attributable to the limited within-class diversity of the cat subset relative to
dog and wild. The dog class remains the most challenging for both models, consistent
with its higher intra-class variability.

\begin{table}[ht]
\centering
\small
\caption{Results on LSUN $256{\times}256$ (15k samples per class).}
\label{tab:lsun}
\resizebox{\textwidth}{!}{%
\begin{tabular}{lcccccc|cccccc|cccccc|cccccc}
\toprule
& \multicolumn{6}{c|}{Global} & \multicolumn{6}{c|}{Bridge} & \multicolumn{6}{c|}{Church} & \multicolumn{6}{c}{Tower} \\
Model & FID$\downarrow$ & KID$\downarrow$ & P$\uparrow$ & R$\uparrow$ & D$\uparrow$ & C$\uparrow$
      & FID$\downarrow$ & KID$\downarrow$ & P$\uparrow$ & R$\uparrow$ & D$\uparrow$ & C$\uparrow$
      & FID$\downarrow$ & KID$\downarrow$ & P$\uparrow$ & R$\uparrow$ & D$\uparrow$ & C$\uparrow$
      & FID$\downarrow$ & KID$\downarrow$ & P$\uparrow$ & R$\uparrow$ & D$\uparrow$ & C$\uparrow$ \\
\midrule
StyleGAN2-ADA
& \textbf{10.85} & \underline{5.39} & \underline{.72} & \textbf{.39} & 0.86 & \textbf{0.81}
& \textbf{19.23} & \underline{9.02} & .62 & \textbf{.30} & 0.63 & \textbf{0.69}
& \textbf{11.99} & \textbf{5.75} & .74 & \textbf{.29} & \underline{1.25} & \textbf{0.81}
& \textbf{15.40} & \underline{6.50} & .68 & \textbf{.33} & 0.70 & \textbf{0.69} \\
DSS-GAN 3-dir
& \underline{12.66} & \textbf{3.45} & \textbf{.79} & \underline{.31} & \textbf{1.08} & \underline{0.75}
& \underline{22.75} & \textbf{6.65} & \textbf{.70} & \underline{.20} & \textbf{0.85} & \underline{0.55}
& \underline{12.11} & \underline{6.35} & \textbf{.81} & \underline{.25} & 1.25 & 0.75
& \underline{18.40} & \textbf{5.44} & \textbf{.76} & .27 & \textbf{0.89} & 0.66 \\
DSS-GAN 2-dir
& 21.79 & 10.30 & .71 & .19 & 0.84 & 0.65
& 35.49 & 20.98 & \underline{.69} & .16 & \underline{0.70} & \underline{0.56}
& 19.54 & 11.00 & \underline{.78} & .14 & 1.11 & \underline{0.75}
& 25.60 & 12.76 & .68 & .22 & 0.76 & \underline{0.65} \\
DSS-GAN 1-dir
& 22.05 & 7.83 & \underline{.72} & .25 & \underline{0.92} & 0.68
& 40.10 & 17.17 & .66 & .19 & 0.58 & 0.54
& 20.52 & 8.21 & \underline{.80} & .21 & \textbf{1.35} & \textbf{0.81}
& 28.70 & 11.31 & \underline{.71} & \underline{.29} & \underline{0.81} & \underline{0.67} \\
\bottomrule
\end{tabular}%
}
\end{table}

\paragraph{LSUN $256{\times}256$.}
The LSUN outdoor dataset~\cite{yu2015lsun} contains three architectural classes (15k samples per
class). As shown in Table~\ref{tab:lsun}, StyleGAN2-ADA achieves lower FID and higher Recall,
while DSS-GAN 3-dir leads in KID (3.45 against 5.39) and in Precision and Density across all
classes. The heterogeneity of the dataset --- a wide range of viewpoints, architectural styles, and
lighting conditions within each class --- favours a model with higher parametric capacity in
covering data variance. KID, being less sensitive to heavy-tailed distributions, consistently
favours DSS-GAN, suggesting higher sample quality near the centre of the real distribution.
Outdoor architectural scenes are dominated by global compositional structure --- perspective,
horizon lines, building symmetry --- rather than by local directional features, making this a
considerably more difficult setting for the DLR mechanism compared to datasets with pronounced
directional texture geometry.

\begin{table}[ht]
\centering
\scriptsize
\caption{Results on FFHQ $256{\times}256$.}
\label{tab:ffhq}
\begin{tabular}{lcccccc}
\toprule
& \multicolumn{6}{c}{Global} \\
Model & FID$\downarrow$ & KID$\downarrow$ & P$\uparrow$ & R$\uparrow$ & D$\uparrow$ & C$\uparrow$ \\
\midrule
StyleGAN2-ADA & \textbf{7.64} & \underline{2.68} & .79 & \textbf{.54} & \underline{1.25} & \underline{.92} \\
DSS-GAN 3-dir & \underline{8.27} & \textbf{2.53} & \textbf{.83} & \underline{.53} & \textbf{1.39} & \textbf{.93} \\
DSS-GAN 1-dir & 8.69 & 3.06 & \underline{.80} & .52 & \underline{1.25} & .90 \\
\bottomrule
\end{tabular}
\end{table}

\paragraph{FFHQ $256{\times}256$.}
The FFHQ dataset~\cite{karras2019stylegan} contains 70k face images without class labels.
Table~\ref{tab:ffhq} shows that DSS-GAN achieves FID 8.27 against 7.64 for StyleGAN2-ADA,
with better KID (2.53 against 2.68), Density (1.39 against 1.23), and Coverage (0.93 against
0.92). Results are comparable; StyleGAN2-ADA holds a slight advantage in FID and Recall, while
DSS-GAN leads in the remaining metrics. It should be noted that StyleGAN2-ADA was evaluated
with the \texttt{paper256} configuration dedicated to this dataset, while DSS-GAN used no
dataset-specific tuning.

\begin{table}[ht]
\centering
\small
\caption{Results on AFHQ $512{\times}512$.}
\label{tab:afhq512}
\resizebox{\textwidth}{!}{%
\begin{tabular}{lcccccc|cccccc|cccccc|cccccc}
\toprule
& \multicolumn{6}{c|}{Global} & \multicolumn{6}{c|}{Cat} & \multicolumn{6}{c|}{Dog} & \multicolumn{6}{c}{Wild} \\
Model & FID$\downarrow$ & KID$\downarrow$ & P$\uparrow$ & R$\uparrow$ & D$\uparrow$ & C$\uparrow$
      & FID$\downarrow$ & KID$\downarrow$ & P$\uparrow$ & R$\uparrow$ & D$\uparrow$ & C$\uparrow$
      & FID$\downarrow$ & KID$\downarrow$ & P$\uparrow$ & R$\uparrow$ & D$\uparrow$ & C$\uparrow$
      & FID$\downarrow$ & KID$\downarrow$ & P$\uparrow$ & R$\uparrow$ & D$\uparrow$ & C$\uparrow$ \\
\midrule
StyleGAN2
& \textbf{11.74} & \textbf{3.06} & \textbf{.86} & .22 & \textbf{1.35} & \textbf{.73}
& \textbf{9.38}  & \textbf{2.55} & .86 & .17 & \textbf{1.69} & \textbf{.85}
& \textbf{27.49} & \textbf{10.40} & \underline{.85} & .25 & \textbf{.80} & \textbf{.51}
& \underline{9.13}  & \textbf{2.22} & \textbf{.82} & .11 & \textbf{1.30} & \textbf{.76} \\

DSS-GAN 1-dir
& 17.77 & 4.91 & .80 & \textbf{.28} & 1.03 & .59
& 12.64 & 5.14 & .90 & .16 & 1.53 & .83
& 39.46 & 25.34 & .77 & \textbf{.43} & .59 & .41
& 12.65 & 3.15  & .71 & \textbf{.23} & .91 & .55 \\

DSS-GAN 3-dir
& \underline{14.44} & \underline{4.11} & \textbf{.86} & \underline{.25} & \underline{1.18} & \underline{.64}
& 12.63 & \underline{4.65} & .91 & .16 & \underline{1.56} & .83
& \underline{36.06} & \underline{22.70} & \textbf{.87} & \textbf{.43} & \textbf{.80} & \underline{.48}
& \textbf{8.81}  & \underline{2.50}  & \underline{.75} & \underline{.19} & \underline{1.22} & \underline{.59} \\

\bottomrule
\end{tabular}%
}
\end{table}

\paragraph{AFHQ $512{\times}512$.}
Table~\ref{tab:afhq512} reports results for DSS-GAN at $512{\times}512$ resolution. The global
FID of 15.65 and Precision of 0.86 for the 3-dir variant are consistent with the proportions
observed at $256{\times}256$, confirming that the architecture scales to higher resolution without
quality degradation. The 3-dir variant outperforms 1-dir across all global metrics --- FID 15.65
against 17.77, KID 4.18 against 4.91, Precision 0.86 against 0.80 --- indicating that the
advantage of directional routing observed at $256{\times}256$ is preserved at higher resolution.
The dog class remains the most challenging (FID 36.06), a consistent pattern across all
resolutions and datasets reflecting the intra-class variability of that category. These results
show that extending the Mamba hierarchy with an additional resolution stage and a StyleGAN2
block at $512{\times}512$ preserves the generative properties of the model without requiring
dataset-specific tuning. It is worth to note that the result for StyleGAN2-ADA is obtained using dedicated tuning for this dataset while tuning for the DSS-GAN is left unchanged.

Generated samples for original datasets are provided in Appendix~\ref{app:examples}.

\begin{table}[ht]
\centering
\scriptsize
\caption{Computational efficiency and memory footprint at $256{\times}256$ resolution, measured
on a single NVIDIA H100 GPU.}
\label{tab:efficiency}
\begin{tabular}{lcccc}
\toprule
\textbf{Model} & \textbf{Parameters} & \textbf{Weight size (MB)} & \textbf{Latency (batch\,1) $\downarrow$} & \textbf{Peak throughput $\uparrow$} \\
\midrule
StyleGAN2-ADA   & 25.0\,M & 24.7 & 9.1\,ms  & \textbf{1451.6\,img/s} ($b{=}128$) \\
DSS-GAN (3-dir) & 7.3\,M  & 21.3 & 15.1\,ms & 280.6\,img/s ($b{=}32$) \\
DSS-GAN (1-dir) & \textbf{4.4\,M}  & \textbf{13.6} & \textbf{8.1\,ms} & 441.6\,img/s ($b{=}32$) \\
\bottomrule
\end{tabular}
\end{table}

\paragraph{Computational Efficiency.}
Table~\ref{tab:efficiency} summarises the inference speed and memory footprint of two DSS-GAN
variants compared to StyleGAN2-ADA at $256{\times}256$ resolution, measured on a single NVIDIA
H100 GPU. Both DSS-GAN variants are substantially more compact: the 3-dir variant requires
7.3M parameters, while the 1-dir variant uses only 4.4M --- an 82\% reduction relative to
StyleGAN2-ADA (25.0M). Despite the sequential nature of state space models, both variants
maintain competitive inference speeds. The 1-dir variant achieves 8.1\,ms latency at batch size 1
(${\approx}122$ FPS), outperforming StyleGAN2-ADA (9.1\,ms) in single-sample generation, and
scales to 441.6\,img/s at $b{=}32$. StyleGAN2-ADA achieves higher peak throughput at large
batches due to fully parallelised convolutions; DSS-GAN provides strong structural control and
parameter efficiency while remaining competitive for latency-critical applications.

\section{Conclusion}

\subsection{Summary}

We presented DSS-GAN, the first generative adversarial network to employ Mamba as a
hierarchical generator backbone for noise-to-image synthesis, operating directly from a latent
vector without any spatial input. The generator processes the feature map through a sequence of
DLR blocks across multiple resolution stages, establishing global spatial coherence via the
SSM recurrence. The architecture was evaluated at resolutions up to $512{\times}512$.

The final resolution stage uses a StyleGAN2-inspired convolutional refinement block rather than
an additional Mamba stage. Each additional SSM stage quadruples the token count, making
high-resolution Mamba processing increasingly costly; replacing it with a convolutional block
balances global coherence with local high-frequency detail that convolutional layers handle more
efficiently. Ablation results confirm that this boundary placement is critical: removing the
convolutional block increases precision but substantially reduces recall, while extending it to
lower resolutions progressively displaces the latent routing responsible for output diversity.

The central contribution is Directional Latent Routing (DLR), a conditioning mechanism that
decomposes the latent vector into direction-specific subvectors, each jointly projected with a
class embedding to produce a feature-wise affine modulation of the corresponding Mamba scan.
Unlike conventional class conditioning schemes that inject a single global signal, DLR couples
class identity and latent structure along distinct spatial axes of the feature map, applied
consistently across all generative scales. The direction weighting network learns non-uniform,
resolution-dependent routing from a uniform initialisation, concentrating class-aware conditioning
precisely at the scales where it has the greatest perceptual impact. Perceptual sensitivity analysis
(LPIPS) confirms that class variations have the strongest impact at low resolutions
($8{\times}8$, $16{\times}16$), consistent with the routing weights learned by the direction
weighting network. Per-direction feature map visualisations further show that individual scan
directions develop spatially distinct activation patterns --- row, column, and diagonal scans
specialise to horizontally, vertically, and diagonally oriented structures respectively --- and that
this specialisation is class-specific, with different classes inducing different activation geometries
across the same set of directions.

An intra-block $180^\circ$ rotation mechanism is introduced to improve gradient stability during
training. Applied and inverted within a single forward pass, it exposes the SSM recurrence to
reversed gradient paths without altering the output spatial layout or introducing additional
parameters. This technique is independent of the DLR mechanism and is applicable to any
unidirectional Mamba architecture operating on spatially structured data.

Experimental results show that the optimal number of scan directions is a property of the
dataset geometry rather than a hyperparameter to maximise --- datasets with pronounced
directional texture structure benefit from multiple directions, while isotropic or rectilinearly
dominated data favour fewer. The importance of scan direction count increases with resolution:
at $128{\times}128$ a single direction can maintain competitive results on several datasets, while
at $256{\times}256$ the performance gap between 1-dir and 3-dir variants widens, reflecting the
larger token sequences over which long-range directional dependencies must be maintained.

DSS-GAN achieves results comparable to or better than StyleGAN2-ADA across all evaluated
datasets, with consistent improvements in KID, Precision, and Density. KID measures sample
quality near the centre of the real distribution and is less sensitive to outliers than FID;
Precision and Density reflect how tightly generated samples concentrate around the real data
manifold. The consistent advantage in these metrics suggests that the hierarchical Mamba
backbone, conditioned via DLR, produces samples that are structurally closer to real data rather
than merely covering the distribution broadly. This comes with an 82\% reduction in parameter count relative to StyleGAN2-ADA, with the
1-dir variant achieving lower single-sample inference latency than the baseline.

Direct comparison with other generative models is limited by differences in experimental
setup: most published results on FFHQ, AFHQ, and LSUN use unconditional training, adaptive
augmentation, or iterative inference. As a point of reference, f-DM~\cite{gu2022fdm}, a
multi-stage diffusion model using 250 denoising steps per sample, reports FID 10.8 on FFHQ
256$\times$256 and FID 5.6--6.4 on AFHQ 256$\times$256. DSS-GAN 3-dir achieves FID 8.27
on FFHQ in a single forward pass, with Precision 0.83 against 0.74 for the best f-DM variant.
On AFHQ, f-DM achieves lower FID (5.6--6.4 vs.\ 10.29), but DSS-GAN reaches Precision 0.88
and Density 1.33 against 0.76--0.81 and 0.48--0.53 for f-DM --- indicating that DSS-GAN
samples concentrate more tightly around the real data manifold despite the FID gap. On LSUN
Church 256$\times$256, f-DM reports FID 8.0--8.2 on the full unconditional dataset; DSS-GAN
is trained conditionally on three architectural classes (bridge, church, tower) and achieves FID 12.11 on the church class, with Precision 0.81 and Density 1.25.
\subsection{Limitations}

Although DSS-GAN achieves competitive results across multiple datasets, several limitations
should be noted. The sequential nature of the SSM recurrence limits peak inference throughput at large batch
sizes relative to fully parallelised convolutional architectures, as reflected in
Table~\ref{tab:efficiency}; this is an inherent property of state space models rather than a
specific design choice.

At higher resolutions, Mamba block capacity (\textit{dstate}, \textit{expand factor}) is progressively reduced to accommodate the quadratically growing token count (Appendix~\ref{app:training}), which may account for part of the performance gap relative to StyleGAN2-ADA at 512×512; at 128×128, where full Mamba capacity is retained, DSS-GAN achieves the strongest relative improvements over StyleGAN2-ADA across all evaluated datasets.

Finally, although the DLR conditioning mechanism is geometry-agnostic in principle and could
accommodate non-standard spatial structures --- including hexagonal, or irregular
grids --- by defining appropriate scan permutations, the CNN-based discriminator is tied to a
regular 2D grid, which restricts the current architecture to standard image domains and would
require replacement to exploit this flexibility in practice.

\subsection{Outlook}

Several directions remain open for future work. The DLR mechanism is architecture-agnostic in
its conditioning principle and could be applied to any sequence-based architecture operating on
spatially structured data --- including discriminative models such as Mamba-based classifiers and
segmentation networks, as well as generative architectures such as diffusion models with Mamba
backbones, where direction-specific latent routing may provide a lightweight alternative to
cross-attention for class and structure conditioning.

Incorporating a Mamba-based discriminator conditioned via DLR is a natural extension not
explored in the present work, where the standard StyleGAN2-ADA discriminator was retained to
isolate the generator contribution. A direction-aware discriminator could provide richer
feedback on the spatial consistency of generated samples along individual scan axes.

At the architectural level, combining DSS-GAN with adaptive discriminator
augmentation~\cite{karras2020ada} was deliberately excluded here to isolate the generator
contribution; applying ADA may further improve results in low-data regimes. Scaling the
architecture beyond $512{\times}512$ by extending the Mamba hierarchy with additional
resolution stages is a direct extension, as results at $512{\times}512$ suggest the model scales
without architectural modification. The directional scanning principle extends naturally to
three-dimensional data by introducing scan directions transverse to the feature map axes ---
corresponding, for instance, to axial, coronal, and sagittal orientations in volumetric medical
imaging --- without modification to the DLR routing mechanism. A systematic grid search over
training hyperparameters --- in particular $R_1$ regularisation strength, learning rate schedule,
and latent space dimensionality --- could be expected to yield further improvements on individual
datasets.

The choice of scan directions examined in this work --- row, column, and diagonal --- covers
the principal spatial axes of two-dimensional feature maps, but the DLR framework places no
constraint on the form of the scan permutation $\pi_k$. Directions designed to capture
specific spatial statistics of the target domain are a natural extension: space-filling curves
such as the Hilbert curve preserve locality and could improve conditioning on datasets with
strong local texture dependencies; Archimedean spiral scans traverse the feature map from
centre outward, potentially capturing radially symmetric structure present in face and
object datasets; Morton (Z-order) curves provide a multi-scale locality-preserving
traversal that may be better suited to hierarchical spatial structure than row or column
scans. Whether a given scan direction is beneficial depends, as shown in Section ~\ref{sec:benchmark}, on
the geometric structure of the underlying data --- direction selection could therefore be
treated as a dataset-specific design choice or learned end-to-end via a differentiable
relaxation of the permutation $\pi_k$.

The original motivation for DSS-GAN was calorimeter response simulation~\cite{krause2024calochallenge},
where sensitivity to directional energy deposition patterns is essential and geometry varies across
detector configurations. Applying DSS-GAN in this domain, where the directional structure of
the data is physically meaningful and well-defined, is a natural next step.

\section{Acknowledgments}

This work was completed with resources provided by the Świerk Computing Centre at the National Centre for Nuclear Research.
We gratefully acknowledge Polish high performance computing infrastructure PLGrid (HPC Centre: ACK Cyfronet AGH) for providing
computer facilities and support within computational grant no. PLG/2024/017403.

\bibliographystyle{plain}
\bibliography{dssgan}

\appendix
\section{Examples}
\label{app:examples}

\newcommand{\figsamplew}{0.35\linewidth}
\newcommand{\figw}{0.45\linewidth}

All samples \textbf{in this appendix section} were generated using truncation $\psi=0.85$, i.e.\ $\mathbf{z} = \psi \cdot \boldsymbol{\epsilon}$, $\boldsymbol{\epsilon} \sim \mathcal{N}(\mathbf{0}, \mathbf{I})$, applied to the input noise prior to inference.

\begin{figure}[H]
    \centering
    \includegraphics[width=0.85\linewidth]{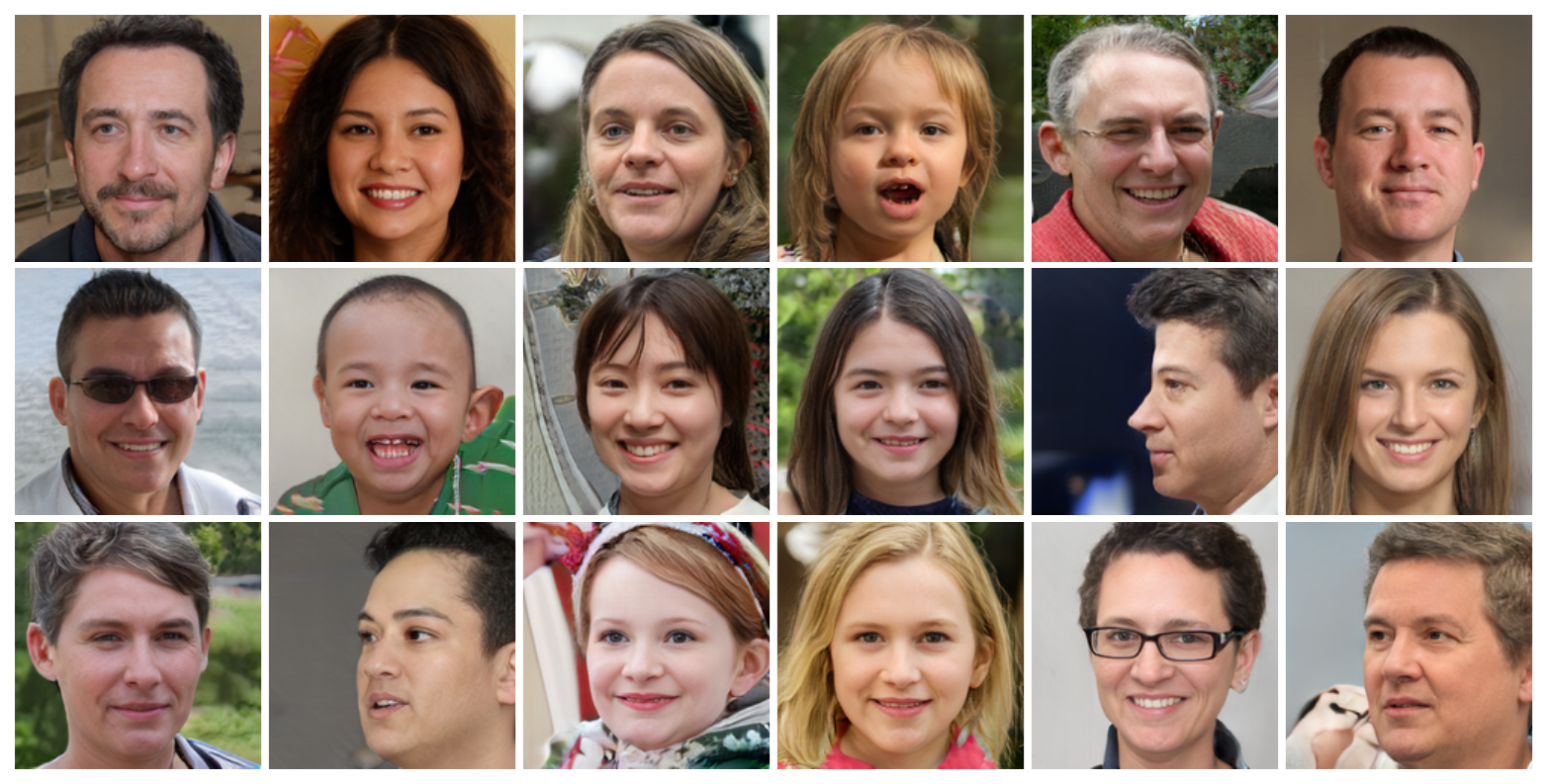}
    \caption{Generated samples, FFHQ 256$\times$256.}
    \label{fig:samples_ffhq_256}
\end{figure}

\begin{figure}[H]
    \centering
    \begin{minipage}{\figsamplew}
        \centering
        \includegraphics[width=\linewidth]{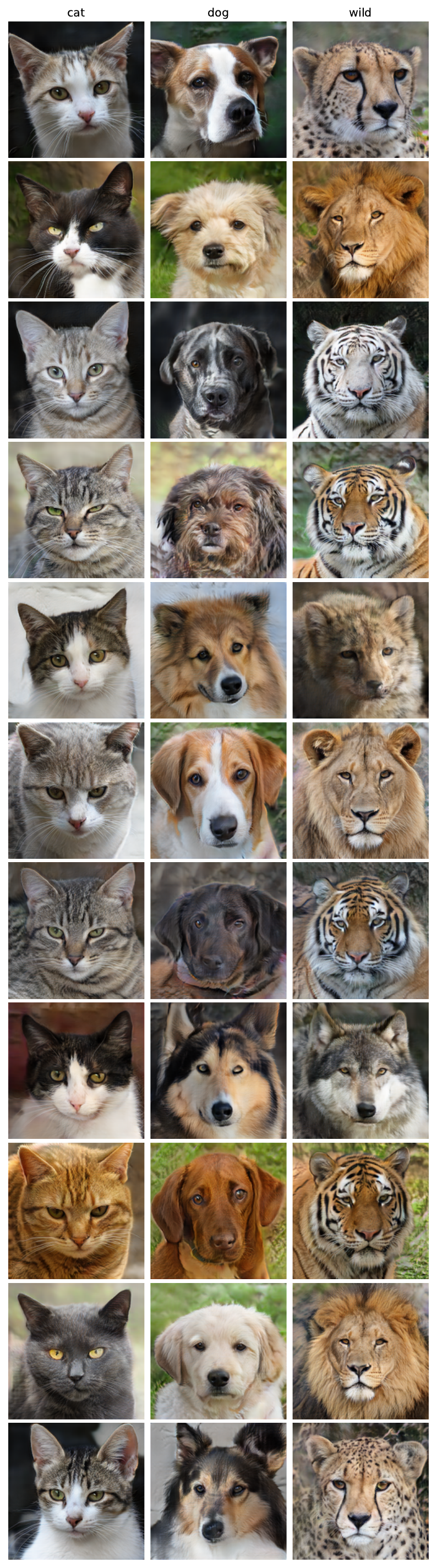}
        \caption{Generated samples, AFHQ 256$\times$256. Same latent space in each row.}
        \label{fig:samples_256}
    \end{minipage}\hfill
    \begin{minipage}{\figsamplew}
        \centering
        \includegraphics[width=\linewidth]{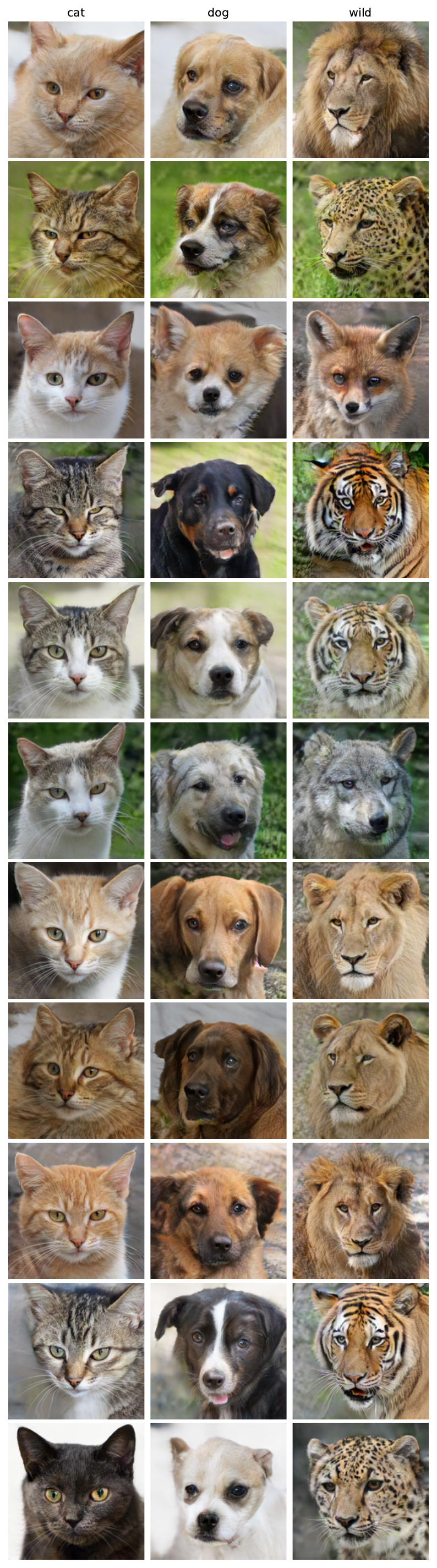}
        \caption{Generated samples, AFHQ 512$\times$512. Same latent space in each row.}
        \label{fig:samples_512}
    \end{minipage}
\end{figure}

\section{Class and Latent Conditioning - LSUN outdoor}
\label{app:lsun_conditioning}

\begin{figure}[H]
    \centering
    \includegraphics[width=\linewidth]{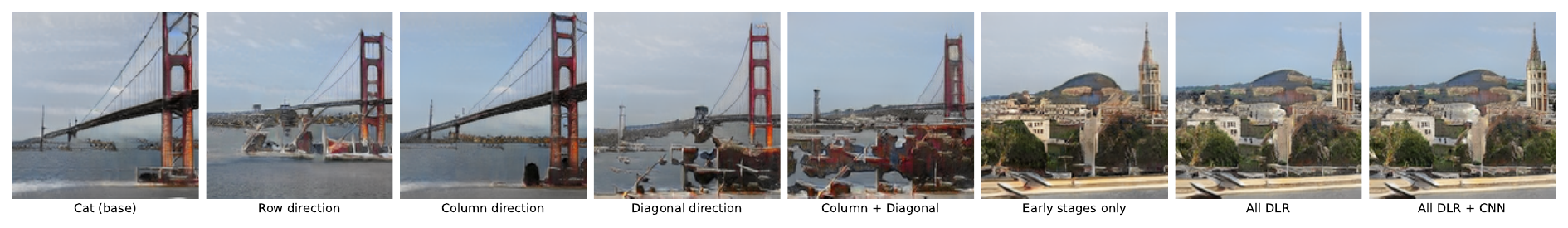}
    \caption{Class embedding override: bridge $\rightarrow$ church. Starting from a base cat image, DLR class embeddings are progressively replaced with church embeddings across individual directions (row, column, diagonal), combinations, early stages only (8$\times$8, 16$\times$16), all DLR directions, and all DLR with CNN refinement block.}
    \label{fig:class_override_bridge_church}
\end{figure}

\begin{figure}[H]
    \centering
    \includegraphics[width=\linewidth]{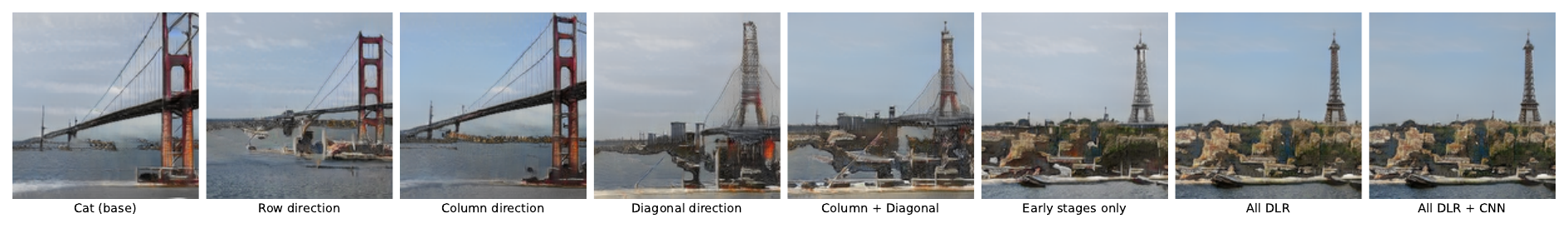}
    \caption{Class embedding override: bridge $\rightarrow$ tower with same protocol as Fig.~\ref{fig:class_override_bridge_church}.}
    \label{fig:class_override_cat_wild}
\end{figure}

\begin{figure}[H]
    \centering
    \includegraphics[width=\linewidth]{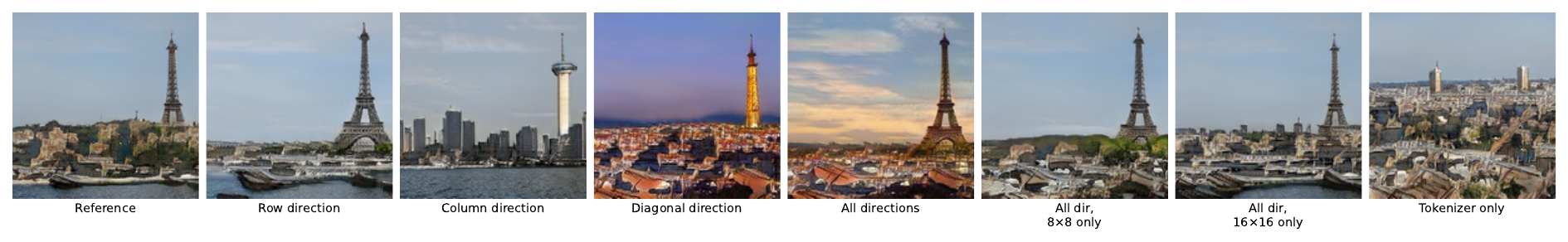}
    \caption{Swapping directional subvectors $z^k_{dir}$ between samples in LSUN dataset. Each column replaces the values of one direction or block stage while keeping all other components of latent fixed. From left: reference, row, column, diagonal, all directions combined, all directions in DLR early-stage only (8×8, 16×16), and tokenizer $\mathbf{z}_\mathrm{base}$ swap.}

\end{figure}

\begin{figure}[h]
    \centering
    \begin{minipage}{\figw}
        \centering
        \small(a)\\[-2pt]
        \includegraphics[width=\linewidth]{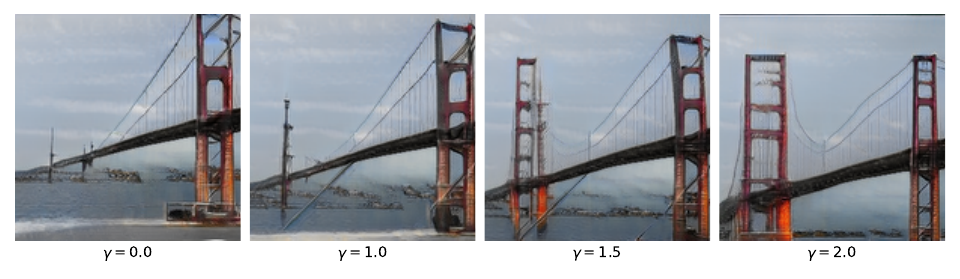}
    \end{minipage}%
    \begin{minipage}{\figw}
        \centering
        \small(b)\\[-2pt]
        \includegraphics[width=\linewidth]{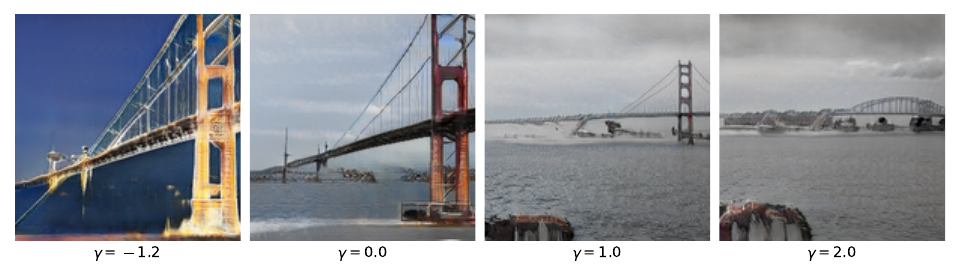}
    \end{minipage}\\[2pt]
    \begin{minipage}{\figw}
        \centering
        \small(c)\\[-2pt]
        \includegraphics[width=\linewidth]{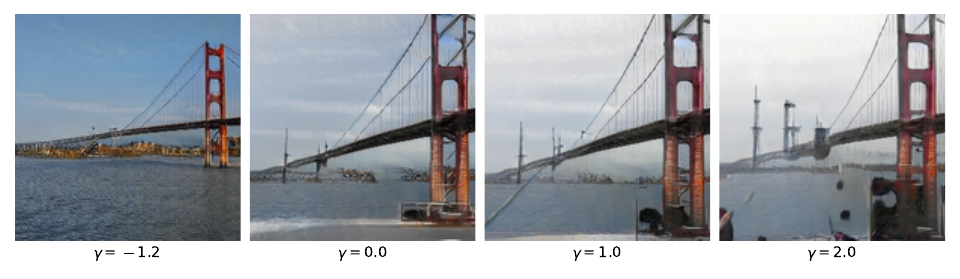}
    \end{minipage}%
    \begin{minipage}{\figw}
        \centering
        \small(d)\\[-2pt]
        \includegraphics[width=\linewidth]{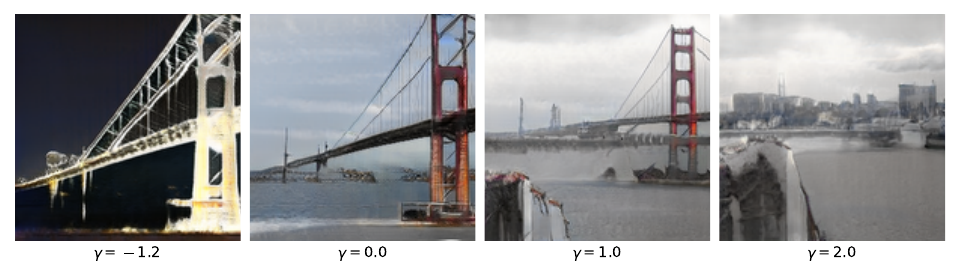}
    \end{minipage}
    \caption{Latent space perturbation analysis (bridge).
    (a)~Tokenizer $\mathbf{z}_\mathrm{tok}$;
    (b)~column direction $\mathbf{z}_\mathrm{dir}^\mathrm{col}$;
    (c)~diagonal direction $\mathbf{z}_\mathrm{dir}^\mathrm{diag}$;
    (d)~combined column and diagonal.
    Each panel shows the effect of scaling perturbation magnitude~$\gamma$.}
\end{figure}

\begin{figure}[H]
    \centering
    \begin{minipage}{\figw}
        \centering
        \small(a)\\[-2pt]
        \includegraphics[width=\linewidth]{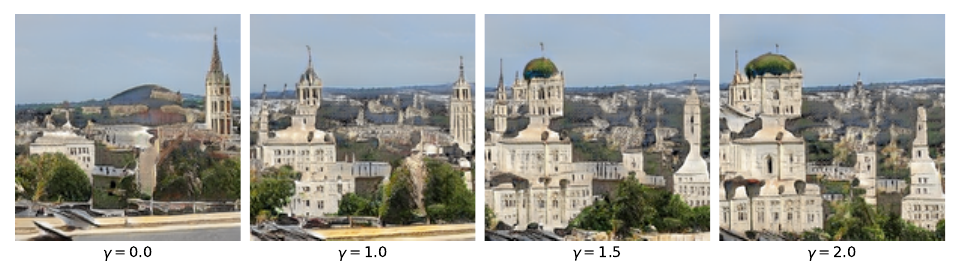}
    \end{minipage}%
    \begin{minipage}{\figw}
        \centering
        \small(b)\\[-2pt]
        \includegraphics[width=\linewidth]{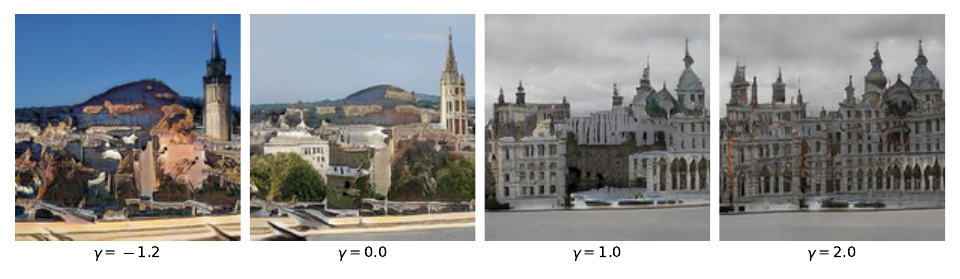}
    \end{minipage}\\[2pt]
    \begin{minipage}{\figw}
        \centering
        \small(c)\\[-2pt]
        \includegraphics[width=\linewidth]{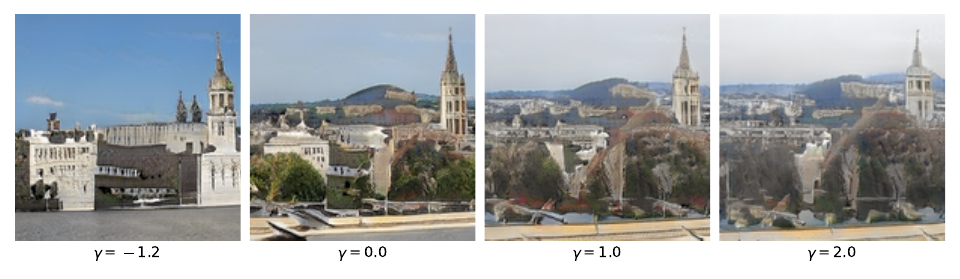}
    \end{minipage}%
    \begin{minipage}{\figw}
        \centering
        \small(d)\\[-2pt]
        \includegraphics[width=\linewidth]{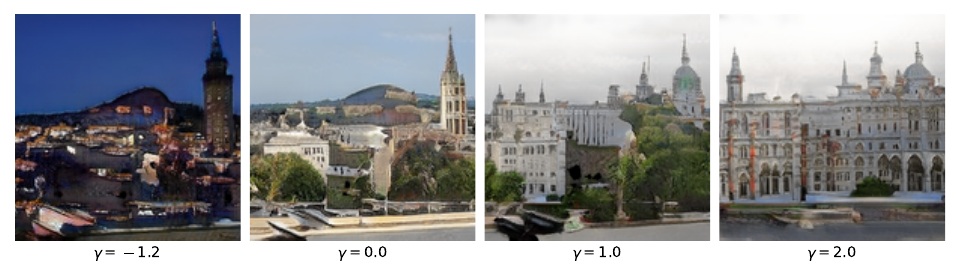}
    \end{minipage}
    \caption{Latent space perturbation analysis (church outdoor).
    (a)~Tokenizer $\mathbf{z}_\mathrm{tok}$;
    (b)~column direction $\mathbf{z}_\mathrm{dir}^\mathrm{col}$;
    (c)~diagonal direction $\mathbf{z}_\mathrm{dir}^\mathrm{diag}$;
    (d)~combined column and diagonal.
    Each panel shows the effect of scaling perturbation magnitude~$\gamma$.}
\end{figure}

\begin{figure}[H]
    \centering
    \begin{minipage}{\figw}
        \centering
        \small(a)\\[-2pt]
        \includegraphics[width=\linewidth]{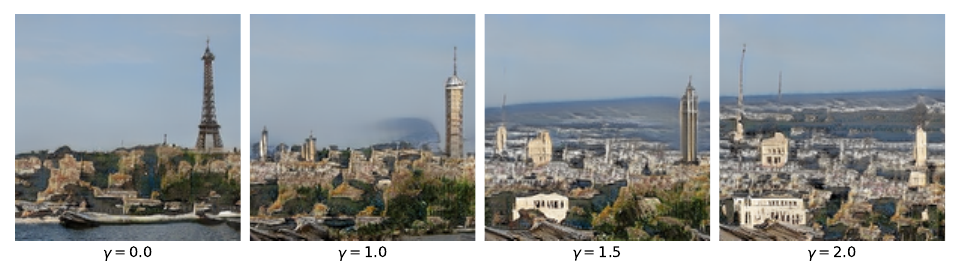}
    \end{minipage}%
    \begin{minipage}{\figw}
        \centering
        \small(b)\\[-2pt]
        \includegraphics[width=\linewidth]{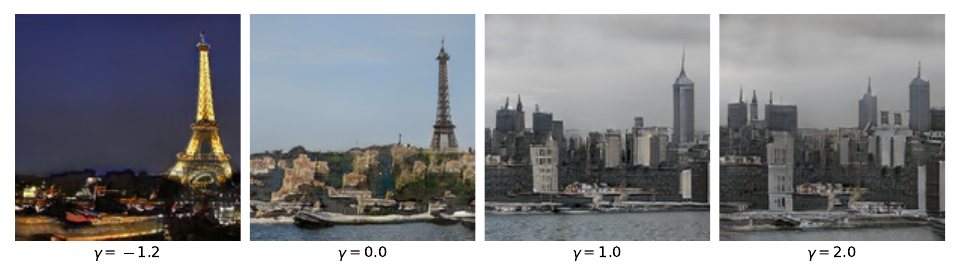}
    \end{minipage}\\[2pt]
    \begin{minipage}{\figw}
        \centering
        \small(c)\\[-2pt]
        \includegraphics[width=\linewidth]{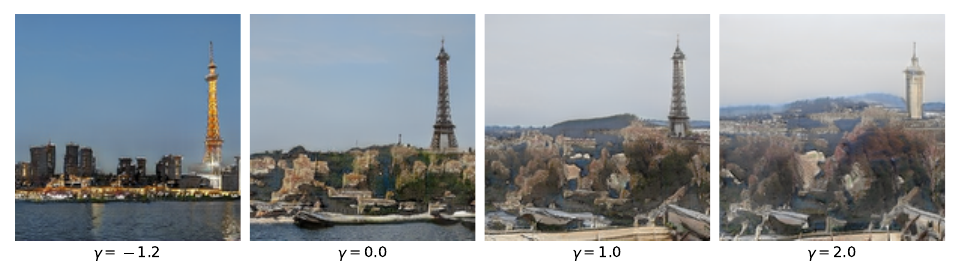}
    \end{minipage}%
    \begin{minipage}{\figw}
        \centering
        \small(d)\\[-2pt]
        \includegraphics[width=\linewidth]{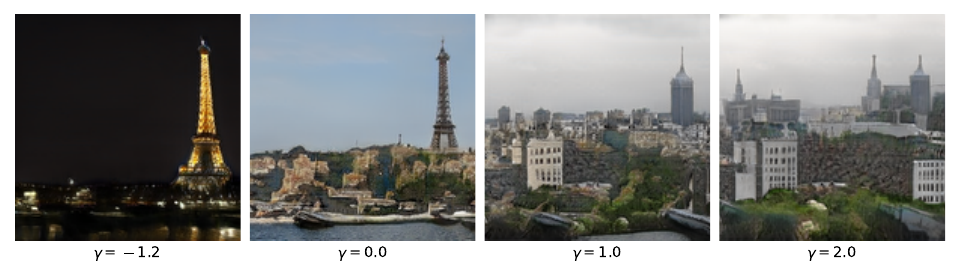}
    \end{minipage}
    \caption{Latent space perturbation analysis (tower).
    (a)~Tokenizer $\mathbf{z}_\mathrm{tok}$;
    (b)~column direction $\mathbf{z}_\mathrm{dir}^\mathrm{col}$;
    (c)~diagonal direction $\mathbf{z}_\mathrm{dir}^\mathrm{diag}$;
    (d)~combined column and diagonal.
    Each panel shows the effect of scaling perturbation magnitude~$\gamma$.}
    \label{fig:noise_perturb_tower}
\end{figure}

\section{Scan Direction Diversity}
\label{app:direction_diversity}

We compare two configurations of $K{=}3$ directions trained on AFHQ
$256{\times}256$ under identical hyperparameters: a baseline in which all three directions
are set to the same row scan, and the standard configuration used throughout the paper
consisting of a horizontal row scan, a vertical column scan, and a diagonal scan.
The two configurations differ only in whether the scan directions provide geometrically
distinct traversal axes of the feature map.

In the same-direction case, each of the $K$ latent subvectors $\mathbf{z}^k_\mathrm{dir}$
conditions a scan that serialises the spatial grid identically: the $K$ directions carry no
independent spatial information. In the distinct case, each direction imposes a different
ordering of spatial positions, exposing the SSM recurrence to different relational structures
and giving the routing network a meaningful basis for specialisation.

\begin{table}[ht]
\centering
\small
\caption{Generation quality comparison between identical and distinct scan direction configurations (AFHQ $256{\times}256$). Best checkpoint per run reported.}
\label{tab:appendix_c}
\resizebox{\textwidth}{!}{%
\begin{tabular}{lcccccc|cccccc|cccccc|cccccc}
\toprule
& \multicolumn{6}{c|}{Global} & \multicolumn{6}{c|}{Cat} & \multicolumn{6}{c|}{Dog} & \multicolumn{6}{c}{Wild} \\
Model
  & FID$\downarrow$ & KID$\downarrow$ & P$\uparrow$ & R$\uparrow$ & D$\uparrow$ & C$\uparrow$
  & FID$\downarrow$ & KID$\downarrow$ & P$\uparrow$ & R$\uparrow$ & D$\uparrow$ & C$\uparrow$
  & FID$\downarrow$ & KID$\downarrow$ & P$\uparrow$ & R$\uparrow$ & D$\uparrow$ & C$\uparrow$
  & FID$\downarrow$ & KID$\downarrow$ & P$\uparrow$ & R$\uparrow$ & D$\uparrow$ & C$\uparrow$ \\
\midrule
DSS-GAN $3{\times}$ same dir
  & 19.20 & 4.16 & \textbf{.894} & .217 & \textbf{1.457} & .569
  & 13.97 & 5.58 & \textbf{.959} & .074 & \textbf{1.750} & .820
  & 38.26 & 21.33 & .899 & .315 & .907 & .463
  & 20.35 & 8.20 & .814 & .133 & 1.090 & .435 \\
DSS-GAN 3-dir
  & \textbf{10.29} & \textbf{2.39} & .879 & \textbf{.356} & 1.327 & \textbf{.743}
  & \textbf{7.88}  & \textbf{2.74} & .903 & \textbf{.293} & 1.652 & \textbf{.872}
  & \textbf{25.15} & \textbf{12.86} & \textbf{.868} & \textbf{.548} & \textbf{.805} & \textbf{.601}
  & \textbf{6.08}  & \textbf{1.35} & \textbf{.877} & \textbf{.150} & \textbf{1.624} & \textbf{.765} \\
\bottomrule
\end{tabular}%
}
\end{table}

\begin{figure}[ht]
\centering
\includegraphics[width=\linewidth]{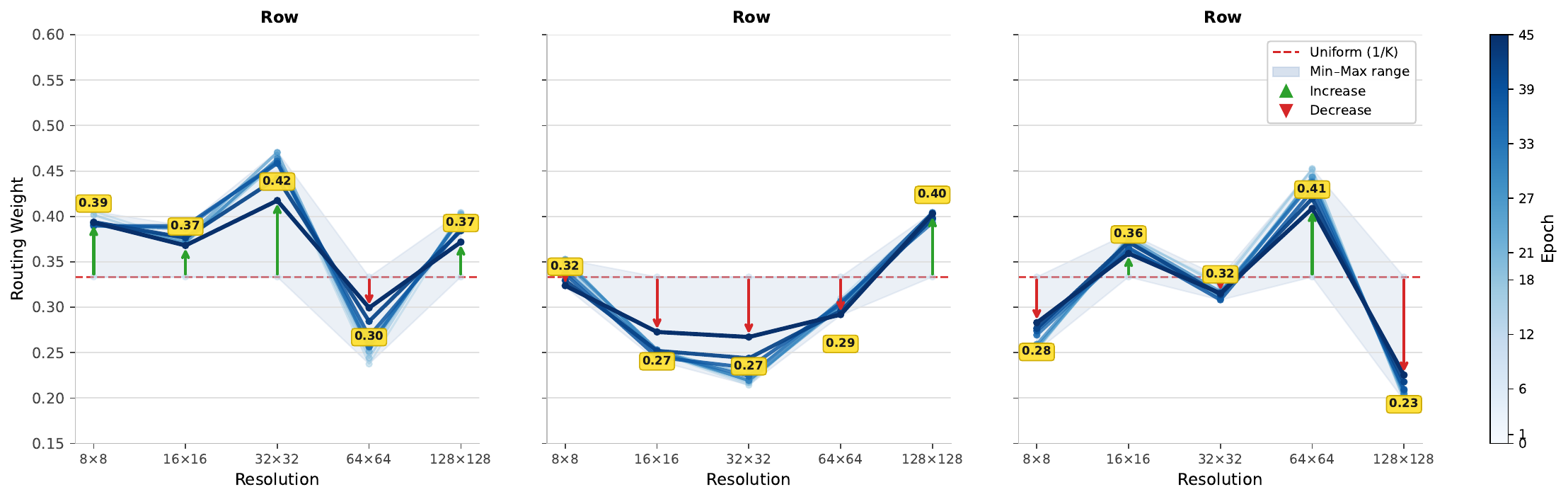}
\caption{Per-direction routing weights across DLR blocks during training
(AFHQ $256{\times}256$), for the \emph{same-direction} ($3{\times}$ row) configuration.
Each line corresponds to a training snapshot; colour indicates epoch (light\,=\,early,
dark\,=\,late). The dashed red line marks the uniform weight $1/K{=}0.33$.
Arrows indicate the direction of change from the first to the last snapshot.
With identical directions, routing weights remain noisy throughout training
and show no convergence toward a stable resolution-dependent pattern,
in contrast to the distinct-direction case (Figure~\ref{fig:routing_evolution}).}
\label{fig:appendix_dirweights}
\end{figure}

\paragraph{Direction specialisation.}
Figure~\ref{fig:appendix_dirweights} shows how routing weights evolve across DLR blocks
during training for the distinct-direction configuration.

Figure~\ref{fig:appendix_dirweights} shows the routing weight profiles for the
same-direction configuration. In contrast to the distinct-direction case
(Figure~\ref{fig:routing_evolution}), where each block develops a clear directional
preference --- the row scan dominant at $8{\times}8$, the column scan peaking at
$16{\times}16$ (0.54), the diagonal scan at $32{\times}32$ (0.43) --- the
same-direction profiles show no structured resolution-dependent routing.
The weights are noisy throughout training and do not converge: the min--max band
is wide relative to the mean, and the final values at different resolutions show
no consistent directional preference. Unlike the distinct-direction case, where
the highest resolution stages converge toward uniformity after early specialisation,
here the weights fluctuate around $1/K$ from the start, with no block ever
becoming dominant.
This is the expected behaviour: because all three scans traverse the feature map in
the same order, the routing network has no geometric basis to assign different weights
to different slots, and the direction weighting mechanism provides no additional
conditioning signal beyond what a single direction would deliver.

This contrasts sharply with Figure ~\ref{fig:routing_evolution}, where the lowest resolution stages show the
strongest specialisation and the highest resolution stages converge toward uniformity ---
a structured, resolution-dependent routing pattern that emerges precisely because each
direction imposes a geometrically distinct traversal axis.

\begin{figure}[ht]
\centering
\includegraphics[width=\linewidth]{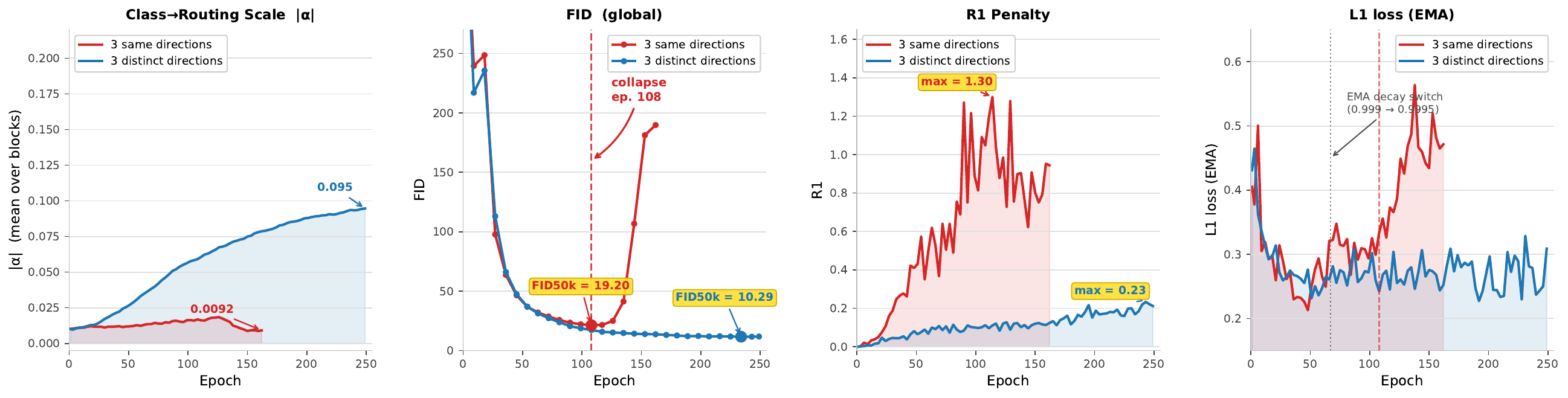}
\caption{Routing diagnostics for $3{\times}$ same (red) vs.\ $3{\times}$ distinct (blue)
directions (AFHQ $256{\times}256$).
From left to right: class-to-routing scale~$|\alpha|$ with EMA decay switch marked
(dotted line); global FID over training epochs with FID at the best
checkpoint annotated; R1 penalty with maximum values annotated; L1 loss (EMA) with EMA
decay switch marked.}
\label{fig:appendix_c_diagnostics}
\end{figure}

\paragraph{Routing collapse.}
Figure~\ref{fig:appendix_c_diagnostics} shows four complementary diagnostic signals for
both configurations.

The class-to-routing scale~$|\alpha|$ is a learned scalar per direction,
initialised near zero, that controls how strongly the class label steers direction
weighting. The direction weights are computed as in Eq.~\ref{eq:softmax},
where the class embedding $\mathbf{e}_y$ is scaled by $|\alpha|$ (Figure ~\ref{fig:dlr}) before being added to
the routing logits.
With distinct directions, $|\alpha|$ grows steadily to 0.095 at the final checkpoint,
indicating that the model has learned to use class identity when selecting which scan axis
to emphasise.
With same directions, $|\alpha|$ remains below 0.010 throughout training --- the routing
network has three times the capacity but no geometric signal to exploit, so the class
routing contribution is never learned enough.

The FID curve reflects the downstream consequence.
With distinct directions, training converges monotonically to
FID\,=\,10.29 at epoch~240.
With same directions, FID plateaus near 21 from epoch~90 and then rises sharply from
epoch~108, eventually reaching 189 --- a pattern consistent with a generator that has
exhausted its capacity for improvement and begins to destabilise.

The R1 penalty measures the gradient norm of the discriminator evaluated at real
samples; elevated R1 indicates that the discriminator loss surface has become steep,
typically in response to mode collapse in the generator.
With distinct directions, R1 remains bounded throughout training, reaching a maximum of
0.23.
With same directions, R1 spikes to 1.30 at epoch~114 --- a 5.7$\times$ increase ---
and remains elevated for the remainder of training, coinciding exactly with the FID
collapse.

The L1 loss (EMA) confirms the same pattern: with distinct directions the
generator loss decreases steadily throughout training; with same directions it rises
sharply after epoch~108, indicating that the generator has lost its ability to produce
outputs that challenge the discriminator.

\paragraph{Summary.}
Providing $K$ identical scan directions prevents the routing mechanism from learning
direction-specific behaviour at every level of the hierarchy: routing weights do not
specialise, the class routing scale $|\alpha|$ does not grow, and the absence of diverse
conditioning eventually destabilises the discriminator gradient landscape, causing training
to collapse.
Geometric diversity of scan directions is therefore a prerequisite for effective DLR ---
not a performance hyperparameter to tune, but a structural requirement for the routing
mechanism to function as intended.

\section{Training Configuration }
\label{app:training}
\begin{table}[H]
\centering
\caption{Training configuration across resolutions. $''$ denotes identical value to the 128$\times$128 baseline.}
\small
\setlength{\tabcolsep}{5pt}
\begin{tabular}{l|l|l|l}
\toprule
\textbf{Hyperparameter} & \textbf{128} & \textbf{256} & \textbf{512} \\
\midrule
\multicolumn{4}{l}{\textit{Optimiser and training}} \\
Adam $(\beta_1,\beta_2)$ & $(0.0,\ 0.99)$ & $''$ & $''$ \\
LR $G$                   & $9{\times}10^{-5}$ & $''$ & $''$ \\
LR $D$                   & $3{\times}10^{-5}$ & $''$ & $''$ \\
Batch size               & 128 & 96 & 48 \\
\midrule
\multicolumn{4}{l}{\textit{Regularisation}} \\
$R_1$ penalty $\gamma$   & 5  & $''$ & $''$ \\
$R_1$ interval           & 4  & $''$ & $''$ \\
Grad clip $G$            & 10 & $''$ & $''$ \\
Grad clip $D$            & 15 & $''$ & $''$ \\
\midrule
\multicolumn{4}{l}{\textit{EMA}} \\
Decay phase 1            & 0.999  & $''$ & $''$ \\
Decay phase 2            & 0.9995 & $''$ & $''$ \\
Switch                   & $10^6$ images & $''$ & $''$ \\
\midrule
\multicolumn{4}{l}{\textit{Latent space}} \\
$D_\text{base}$          & 92  & $''$ & $''$ \\
$D_\text{dir}$           & 20  & 28 & $''$ \\
$D_e$                    & 64  & $''$ & $''$ \\
Routing temp.\ $\tau$    & 1.0 & $''$ & $''$ \\
\midrule
\multicolumn{4}{l}{\textit{Generator channels}} \\
$8{\times}8$ --- $32{\times}32$             & 148 & $''$ & $''$ \\

$64{\times}64$           & 148 & $''$ & 128 \\
$128{\times}128$         & 168 (CNN) & 148 (M) & 98 (M) \\
$256{\times}256$         & —   & 216 (CNN) & 72 (M) \\
$512{\times}512$         & —   & —         & 64 (CNN) \\
\midrule
\multicolumn{4}{l}{\textit{Mamba tokenizer}} \\
Depth                    & 2  & $''$ & $''$ \\
$d_\text{state}$         & 64 & $''$ & $''$ \\
$d_\text{conv}$          & 4  & $''$ & $''$ \\
Expand                   & 2  & $''$ & $''$ \\
\midrule
\multicolumn{4}{l}{\textit{Mamba blocks}} \\
Depth $8{\times}8$                             & 2   & $''$ & $''$ \\
Depth $16{\times}16$+                          & 1   & $''$ & $''$ \\
$d_\text{state}$ $8{\times}8$--$32{\times}32$  & 64  & $''$ & $''$ \\
$d_\text{state}$ $64{\times}64$                & 64  & 48   & $''$ \\
$d_\text{state}$ $128{\times}128$              & —   & 48   & 32 \\
$d_\text{state}$ $256{\times}256$              & —   & —    & 16 \\
$d_\text{conv}$ $8{\times}8$--$32{\times}32$   & 4   & $''$ & $''$ \\
$d_\text{conv}$ $64{\times}64$                 & 3   & $''$ & $''$ \\
$d_\text{conv}$ $128{\times}128$               & —   & 3    & 2 \\
$d_\text{conv}$ $256{\times}256$               & —   & —    & 1 \\
Expand $8{\times}8$--$32{\times}32$            & 2.0 & 1.5  & 1.0 \\
Expand $64{\times}64$                          & 1.5 & 1.0  & $''$ \\
Expand $128{\times}128$                        & —   & 1.0  & $''$ \\
Expand $256{\times}256$                        & —   & —    & 1.0 \\
\midrule
\multicolumn{4}{l}{\textit{Discriminator (StyleGAN2)}} \\
Base channels            & 96  & 64  & 16 \\
Max channels             & 512 & 384 & 256 \\
Minibatch std group      & 8   & $''$ & $''$ \\
\bottomrule
\end{tabular}
\end{table}

\end{document}